\documentclass[journal]{IEEEtran}
\usepackage{multirow} 
\usepackage{booktabs}
\usepackage{makecell}
\usepackage{multicol}
\usepackage{cite}
\usepackage{amsmath} 
\usepackage{lineno}
\usepackage[ruled]{algorithm2e} 
\usepackage{array}
\usepackage{graphicx}
\usepackage{float}
\usepackage{amsthm,amsmath,amssymb}
\usepackage[table,xcdraw]{xcolor}
\graphicspath{{./figures/}}
\usepackage{url}  

\ifCLASSOPTIONcompsoc
  \usepackage[caption=false,font=normalsize,labelfont=sf,textfont=sf]{subfig}
\else
  \usepackage[caption=false,font=footnotesize]{subfig}
\fi

\begin{document}
\title{DSU-Net: Dynamic Snake U-Net for 2-D Seismic First Break Picking}
\author{Hongtao~Wang, Rongyu~Feng, Liangyi~Wu, Mutian~Liu, Yinuo~Cui, Chunxia~Zhang*, Zhenbo~Guo.
\thanks{Corresponding authors: Chunxia~Zhang}
\thanks{H.T. Wang, R.Y. Feng, L.Y. Wu, M.T. Liu, Y.N. Cui, C.X. Zhang, are with the School of Mathematics and Statistics, Xi'an Jiaotong University, Xi'an, Shaanxi, 710049, P.R.China.}
\thanks{Z.B. Guo is with China National Petroleum Corp Bureau of Geophysical Prospecting Inc, Geophysical Technology Research Center Zhuozhou, Hebei, 072750, P.R.China.}
\thanks{The research is supported by the National Natural Science Foundation of China under grant 12371512.}}

\markboth{Journal of \LaTeX\ Class Files,~Vol.~14, No.~8, May~2023}%
{Wang \MakeLowercase{\textit{et al.}}: DSU-Net for First Break Picking}

\maketitle

\begin{abstract}
In seismic exploration, identifying the first break (FB) is a critical component in establishing subsurface velocity models. Various automatic picking techniques based on deep neural networks have been developed to expedite this procedure. The most popular class is using semantic segmentation networks to pick on a shot gather called 2-dimensional (2-D) picking. Generally, 2-D segmentation-based picking methods input an image of a shot gather, and output a binary segmentation map, in which the maximum of each column is the location of FB. However, current designed segmentation networks is difficult to ensure the horizontal continuity of the segmentation. Additionally, FB jumps also exist in some areas, and it is not easy for current networks to detect such jumps. Therefore, it is important to pick as much as possible and ensure horizontal continuity. To alleviate this problem, we propose a novel semantic segmentation network for the 2-D seismic FB picking task, where we introduce the dynamic snake convolution into U-Net and call the new segmentation network dynamic-snake U-Net (DSU-Net). Specifically, we develop original dynamic-snake convolution (DSConv) in CV and propose a novel DSConv module, which can extract the horizontal continuous feature in the shallow feature of the shot gather. Many experiments have shown that DSU-Net demonstrates higher accuracy and robustness than the other 2-D segmentation-based models, achieving state-of-the-art (SOTA) performance in 2-D seismic field surveys. Particularly, it can effectively detect FB jumps and better ensure the horizontal continuity of FB. In addition, the ablation experiment and the anti-noise experiment, respectively, verify the optimal structure of the DSConv module and the robustness of the picking.

\end{abstract}

\IEEEpeerreviewmaketitle

\section{Introduction}
\IEEEPARstart{I}{dentifying} the first break (FB) of P-wave or S-wave in pre-stack seismic gather is a critical challenge in seismic data processing. Precise detection of this first break is essential as it can lead to accurate static correction outcomes, thereby enhancing the effectiveness of subsequent seismic data processing tasks such as velocity analysis and stratigraphic imaging\cite{yilmaz2001seismic}. The conventional manual FB picking method is no longer adequate due to the growing volume of collected seismic data. Consequently, automating the FB picking process has emerged as a key objective in expediting seismic data processing.

In the manual FB picking processing, the analyst regards the first change point, which represents the arrival of the first wave, as the first break time. Therefore, the initial automatic picking method is based on signal processing or constructing higher-order statistics to detect the change point. For instance, the short- and long-time average ratio (STA/LTA) measures the degree of local variation of a time series signal (a single trace in a shot gather)\cite{allen1978automatic,baer1987automatic}. Concretely, the location where the STA/LTA value exceeds the manual threshold for the first time is considered as the FB. Subsequently, more energy ratios and higher-order statistics were defined to detect FB, such as modified energy ratio (MER)\cite{gaci2013use}, modified Coppens' method (MCM)\cite{sabbione2010automatic}, Akaike information criterion (AIC)\cite{takanami1991estimation}, Kurtosis and skewness\cite{Saragiotis2004}, etc. However, the above traditional methods can only solve the picking case of high signal-to-noise ratio (SNR). Thus, Kim et al.\cite{kim2023first} proposed a new picking algorithm based on the differences between multi-window energy ratios (DERs) that minimizes the effects of noise. 
Although a few traditional signal processing-based picking methods can alleviate the influence of noise on picking, when faced with complex noise, such as multi-wave noise, the above method can not eliminate the interference of noise. Therefore, in the era of artificial intelligence (AI), various automatic picking methods based on machine learning (ML) have developed rapidly to solve the picking problem under complex noise.

Unlike signal processing-based methods, ML-based methods acquire the concept of FB from ample data. These methods are also referred to as data-driven picking methods. 
In the earliest study, McCormack et al. \cite{mccormack1993first} used a two-layer neural network classification method to binary classify the 1-dimensional range where FB is located. After the rise of convolutional neural networks (CNN), CNNs were used to binary classify the FB for each 2-dimensional patch on a shot gather\cite{yuan2018seismic,duan2020multitrace}. In addition to neural networks, support vector machine (SVM), a classical ML classification method, was also used for local classification of FB \cite{duan2019multi}. 
After full convolutional networks (FCNs) have achieved great success in semantic segmentation in computer vision \cite{long2015fully}, FCN was introduced into the FB pick task. Tsai et al. \cite{tsai2019automatic} proposed an FCN-based semi-supervised learning picking method to solve the FB picking problem in an end-to-end framework. 
Since then, semantic segmentation-based methods have become the most popular automatic picking method. However, the FB picking task requires high precision, where the absolute error between the automatic picking and the manual picking of each trace must be at most 3ms. 
Under this requirement, the semantic segmentation network used for FB picking should pay more attention to the texture structure of the first arrival wave. Fortunately, U-Net \cite{ronneberger2015u}, as a robust and efficient medical image segmentation network, fulfills this need. Specifically, Hu et al. \cite{hu2019first} utilized U-Net to segment the shot gather into a binary classification map. They then extracted the FB for each trace from the segmentation map by applying a manual threshold. In the same year, Ma et al. \cite{ma2019automatic} also developed a UNet-based picking method. 
After this, much work has been completed on semantic segmentation based on the U-Net architecture, but the focus of improvement varies. 

Since our method falls within the U-Net-based picking framework, we briefly summarize the related works as follows.
First, various U-Net variants are utilized in the FB picking process. Zhang and Sheng \cite{zhang2020first} proposed a Residual Link Nested U-Net Network (RLU-Net) to identify the FB. Subsequently, the Swin Transformer UNet was developed to enhance the generalization of the FB picking network \cite{jiang2023seismic}. Recently, an improved UNet model was proposed and evaluated on an open-source dataset\cite{st2024deep}. This study establishes a strong benchmark model for deep learning-based picking research. Subsequently, Shen et al. \cite{shen2024improved} proposed a more complex U-Net variant named improved U-Net3+ to enhance the picking accuracy of noisy earthquake recordings.
Second, in addition to improving the basic architecture of the segmentation network, some studies focus on enhancing picking performance by optimizing the picking framework. Ozawa \cite{ozawa2024automated} proposed a novel two-stage picking framework. In this framework, a 1-dimensional U-Net is used for the initial pick, followed by a 2-dimensional U-Net, which picks the FB again based on the corrected FB of the 1-dimensional U-Net output. This framework alleviates some of the dilemmas associated with incorrect manual labeling and enhances the augmentation of training samples. Different from this framework, Wang et al. \cite{wang2024msspn} proposed a multi-stage segmentation picking network, in which the picking task in the case of low SNR is decomposed into four sub-missions to simulate manual picking processes. 
Third, U-Net has been extended to higher dimensions to leverage the spatial correlation of seismic traces for improved picking stability. Han et al. \cite{han2021first} utilized a 3-dimensional U-Net to pick the samples arranged by the intersection of the shot line and the receiver line. However, when the observation system of the actual collection area is irregular, it will be very difficult to construct 3-dimensional seismic samples. To alleviate the challenge of constructing data cubes, Wang et al. \cite{wang2024seismic} overcame the constraints of traditional dimensional gathering by suggesting the construction of trace sets into graphs. They utilized a graph neural network (GNN) to pick the FB for each node based on a 1-dimensional UNet variant.
Eventually, some studies incorporated the regional velocity prior to the picking framework to constrain the picking region of U-Net. For instance, Cova et al. \cite{cova2020automated} proposed velocity-based constrained pooling networks and combined them with U-Net to enhance robustness.

This study focuses on the fundamental structure of the segmentation network used for FB picking. To further enhance the precision and accuracy of the FB picking, the segmentation network must incorporate the characteristics of the seismic pre-stack gather, such as the continuity of the FBs in the shot gather. We develop a novel U-Net variant based on the characteristics of gather continuity. This variant is more effective at capturing the continuous texture state of nearby traces. Specifically, we utilize the dynamic snake convolutional network \cite{qi2023dynamic} as a reference, enhance the convolutional module of U-Net, and develop a segmentation network called dynamic snake U-Net (DSU-Net).

The structure of this article is summarized as follows: Section II describes the details of our proposed DSU-Net. Section III shows the performance validation of the proposed method on four field datasets. Finally, Section IV concludes this paper.

\section{Methodology}
This section introduces the fundamental concepts of traditional convolution and dynamic snake convolution. Subsequently, the details of the proposed dynamic snake U-Net will be discussed.

\subsection{Traditional Convolution and Dynamic Snake Convolution}
\begin{figure}[!ht]
    \centering
    \includegraphics[width=0.4\textwidth]{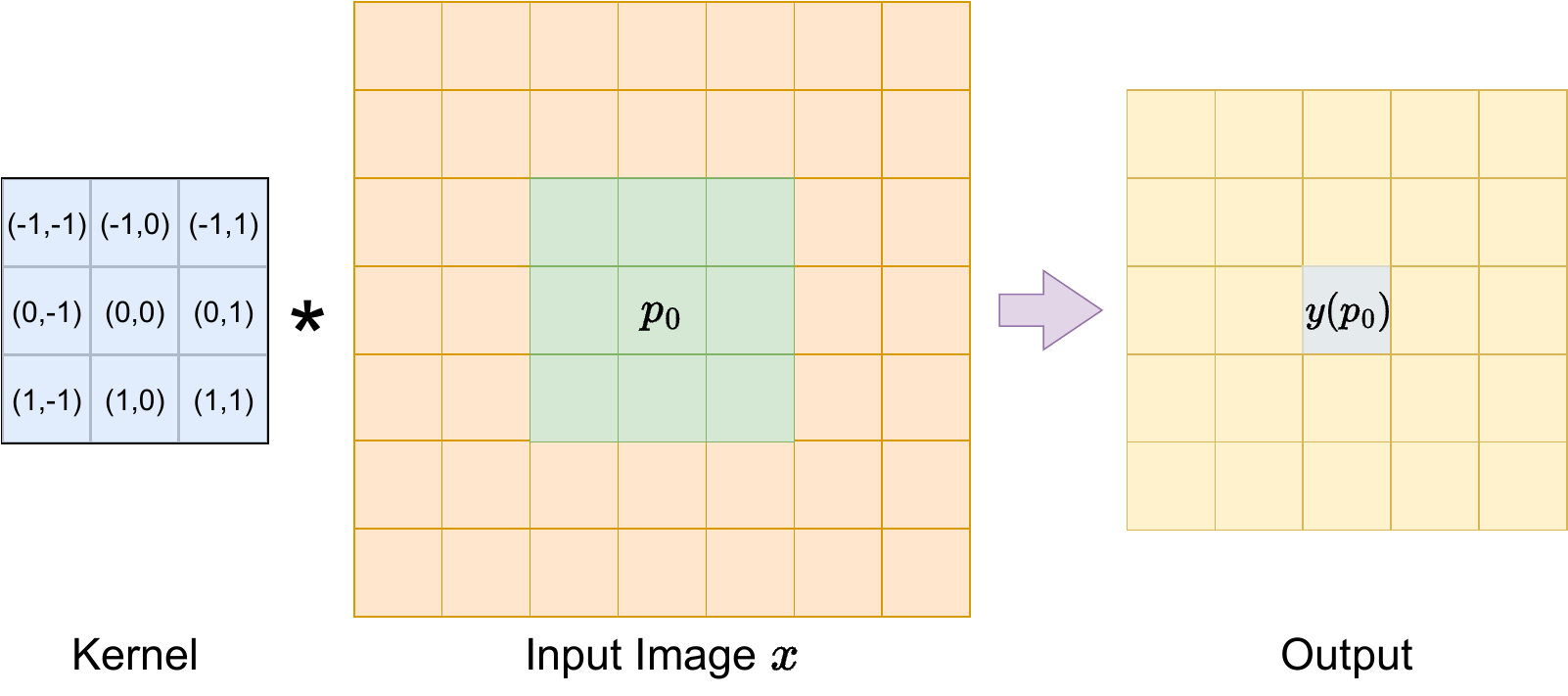}
    \caption{A showcase of the 3$\times$3 convolution operation.}
    \label{fig:tra_conv}
\end{figure}

Convolution operation in CNNs is a mathematical process involving a kernel or filter applied to the input data, where the kernel slides across the input data, performing element-wise multiplication and summing the results to produce a feature map.
For instance, a traditional convolutional computation with the kernel size $= 3\times 3$  is shown in Fig. \ref{fig:tra_conv}.
The coordinate set of the traditional convolution kernel is defined on a $3\times 3$ grid: $$\mathbb{R} \triangleq \left\{(-1,-1),(-1,0),...,(0,1),(1,1)\right\},$$ and $f_w$ is the weight function defined on $\mathbb{R}$, which maps a coordinate to a scale value.
Concretely, the convolutional computation can be defined by:
\begin{equation}
   y(p_0)\triangleq \sum_{p_k\in \mathbb{R}}{f_w(p_k)\ast x(p_0+p_k)},
 \label{eq:conv}
\end{equation}
where $y(\cdot)$ represents the pixel value of the corresponding point in the output image, and $x(\cdot)$ is the mapping of coordinates to pixel values in the input image. $p_0$ refers to the coordinate of the pixel point in the input image that corresponds to the center position of the convolutional kernel as it slides along the input image. 

\noindent
This means that each pixel value in the output image after convolution is determined by the surrounding 3×3 pixel values of the original image at the corresponding coordinates, along with the weights of the convolutional kernel. The fixed correspondence between weights and pixel point coordinates limits the receptive field of the convolutional kernel. For conventional convolutional kernels, enlarging the size of the kernel is necessary to increase the receptive field. However, an excessively large kernel size results in higher computational costs. Moreover, the fixed coordinate correspondence relationship makes the convolutional kernel rigidly extract features from a fixed area, which can be inefficient and insufficient when the features to be extracted have obvious geometric shapes or complex structures.

\begin{figure}[!ht]
    \centering
    \subfloat[]{\includegraphics[width=0.12\textwidth]{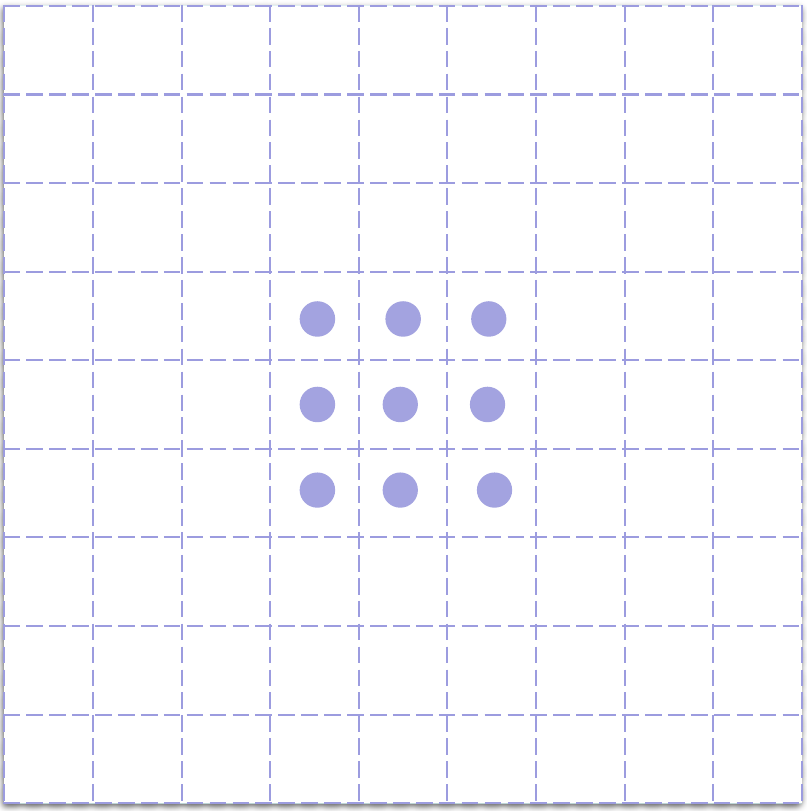}\label{fig:tra_conv2}}
    \subfloat[]{\includegraphics[width=0.12\textwidth]{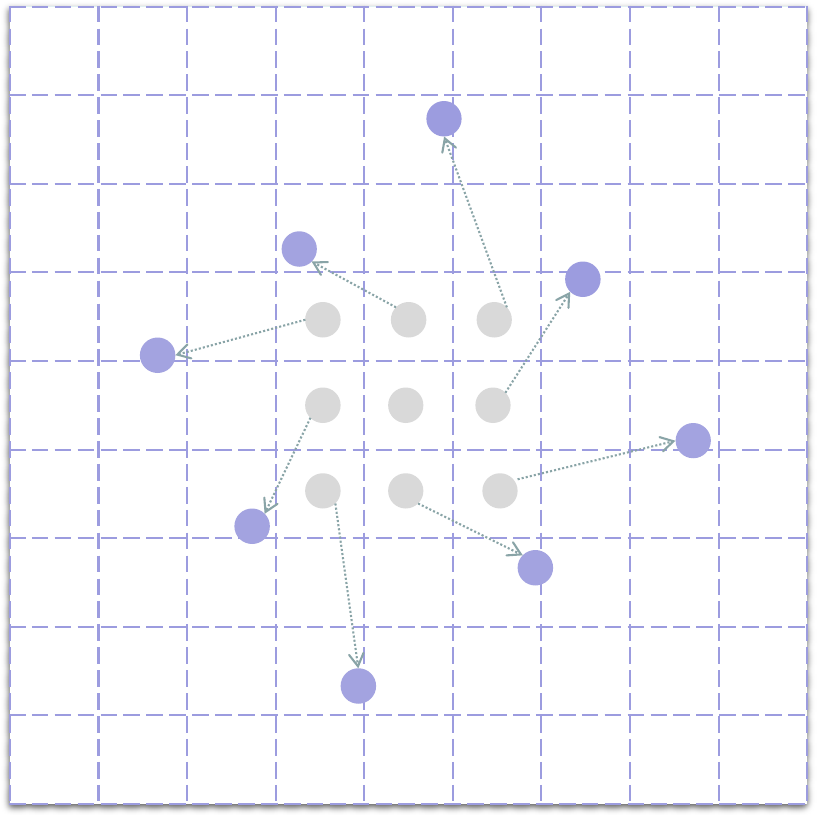}\label{fig:de_conv}}
    \subfloat[]{\includegraphics[width=0.12\textwidth]{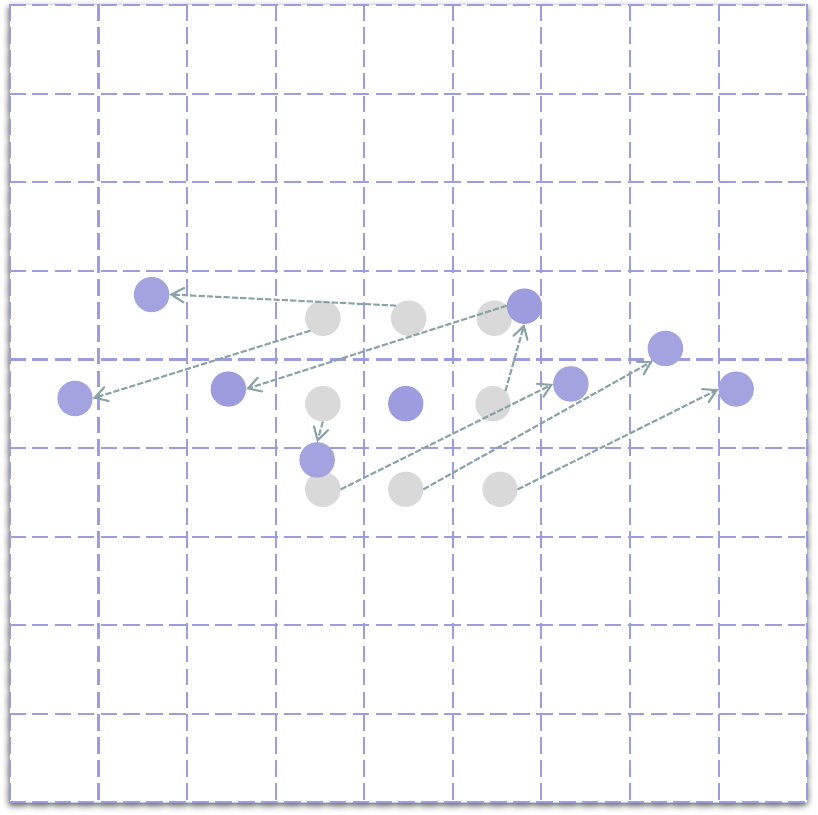}\label{fig:de_conv_x}}
    \subfloat[]{\includegraphics[width=0.12\textwidth]{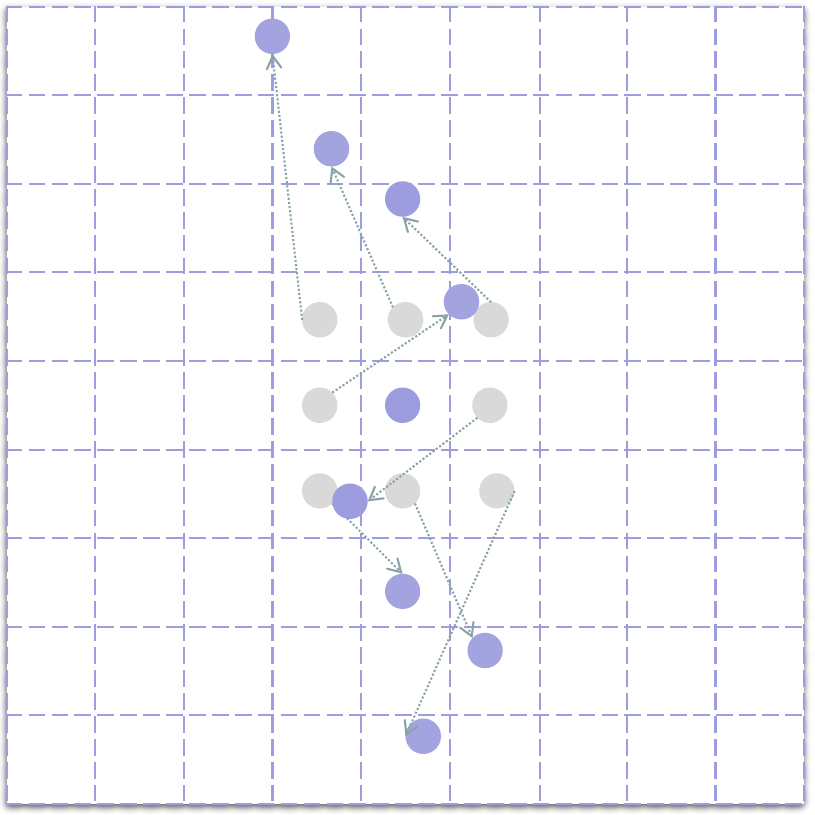}\label{fig:de_conv_y}}
    \caption{Three types of convolution kernels. (a) Traditional convolution kernel. (b) Deformable convolution kernel. (c) X-direction dynamic snake convolution kernel. (d) Y-direction dynamic snake convolution kernel. }
    \label{fig:dsconv}
\end{figure}

To better adapt to complex structures through the design of specific network architectures and modules, researchers have developed deformable convolutional networks based on the design of convolutional kernels\cite{dai2017deformable}. The deformable convolution approach addresses the intrinsic geometric transformations (Fig.~\ref{fig:tra_conv2}) in convolutional neural networks to adapt to structures with varying shapes (Fig.~\ref{fig:de_conv}).
Specifically, the main idea of the deformable convolution is to learn the kernel offset automatically using a neural network. The deformable convolution computation is defined by:
\begin{equation}
   y(p_0)\triangleq \sum_{p_k\in \mathbb{R}}{f_w(p_k)\ast x(p_0+p_k+\Delta p_k)}, 
 \label{eq:de_conv}
\end{equation}
where $\Delta p_k$ represents the learned offset for $k$-th position and $\mathbb{R}$ is the coordinate set of the convolution kernel.
However, the learning direction of deformable offsets cannot be controlled. Thus, we cannot specify that the deformable convolution focus on the tubular structure in the image, which is required for the FB picking task. Drawing inspiration from dynamic snake convolution (DSConv)\cite{qi2023dynamic}, we incorporate DSConv into the U-Net architecture, aiming to enhance the ability of recognizing small tubular features.
Unlike deformable convolution, DSConv fixes the change of one dimension during deformation, i.e., deforming the convolution kernel along the horizontal direction (x-direction) or vertical direction (y-direction), as shown in Fig.~\ref{fig:de_conv_x} and Fig.~\ref{fig:de_conv_y}, respectively. 
For instance, considering a $3\times 3$ convolution kernel and deforming it along the x-direction, the kernel position set for $i$-pixel is defined by: 
\begin{equation}
   \mathbb{R}_i \triangleq \left\{r_{i\pm c}\right\}_{c=0}^{4} = \left\{(x_{i\pm c}, y_{i\pm c})\right\}_{c=0}^{4},
 \label{eq:de_conv2}
\end{equation}
where $c$ represents the horizontal distance from the central pixel. 
The computation of each grid $r_{i\pm c}$ in $\mathbb{R}$ is an accumulation process. In the case of $c$ equal to 2, we first calculate the position of $r_1$, and then shift it again according to $r_1$ to get $r_2$:
\begin{equation}
    \begin{array}{c}
      y_2 = y_1 + \Delta y_1 = y_0 + \Delta y_0 + \Delta y_1;\\
      x_2 = x_1 + 1 = x_0 + 1 + 1 = x_0 + 2,
    \end{array}
 \label{eq:ds_conv_show_case}
\end{equation}
where $\Delta y_1$ and $\Delta y_0$ is the vertical offset for $r_1$ and $r_0$.
Subsequently, the x-direction DSConv for any $c$ can be defined by:
\begin{equation}
    \begin{array}{c}
      r_{i+c} \triangleq (x_i+c, y_i+s^e\ast\sum_{i}^{i+c}\Delta y),\\
      r_{i-c} \triangleq (x_i-c, y_i+s^e\ast\sum_{i-c}^{i}\Delta y),
    \end{array}
 \label{eq:ds_conv_x}
\end{equation}
where $s^e$ represents the extension scope, which controls the research range of the kernel.
\noindent
Similarly, the y-direction DSConv can be defined by:
\begin{equation}
    \begin{array}{c}
      r_{j+c} \triangleq (x_j+s^e\ast\sum_{j}^{j+c}\Delta x, y_j+c),\\
      r_{j-c} \triangleq (x_j+s^e\ast\sum_{j-c}^{j}\Delta x, y_j-c).
    \end{array}
 \label{eq:ds_conv_y}
\end{equation}
Particularly, since the offset is usually a float, and the coordinates are typically in integer form, bilinear interpolation is performed using the nearest four integer coordinate points adjacent to it.

\begin{figure*}[!ht]
    \centering
    \includegraphics[width=\textwidth]{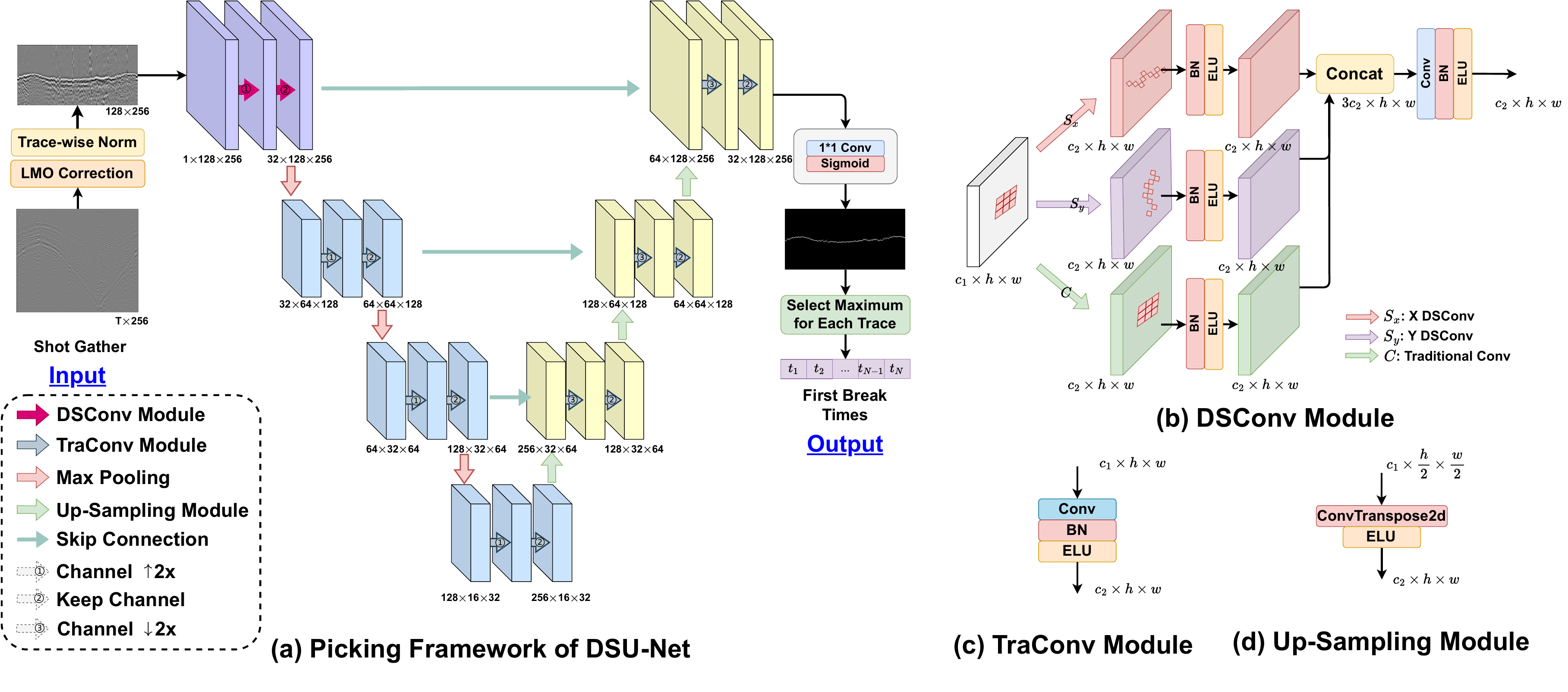}
    \caption{The structure chart of DSU-Net. (a) The basic architecture of DSU-Net. In the first layer of the encoder, we utilize a stack of two DSU modules to deeply extract superficial features. (b) The specific structure of the DSU module. (c) The traditional convolution (TraConv) module. (d) Up-sampling module.}
    \label{fig:DSUNet}
\end{figure*}

\subsection{Dynamic Snake U-Net}
The overall automatic picking based on the DSU-Net involves three steps: preprocessing of seismic gather, semantic segmentation of FB, and post-processing of network outputs, as illustrated in Fig.~\ref{fig:DSUNet}a. During preprocessing, the 2-D shot gather is cropped using LMO correction. Subsequently, we normalize the cropped gather to the range of $[-1, 1]$ and feed the normalized gather into DSU-Net. Then, the DSU-Net outputs a segmentation map of the FB. Finally, the FB can be obtained from the segmentation map using a post-processing method. In the remaining subsection, we will describe each of these three steps separately.

During the propagation of seismic waves, the range of the FB time is determined by the velocity. Thus, to reduce the difficulty of picking, we utilize linear moveout (LMO) correction to narrow the picking range:
\begin{equation}
    t'=t_0+\frac{x}{v},
 \label{eq:LMO}
\end{equation}
where $t'$ represents the corrected static displacement, $x$ is the offset of the corrected trace, $t_0$ is the time when the source-receiver offset tends to zero (i.e., the interception time), and $v$ is the reference velocity. The LMO correction is determined by the hyperparameters $v$ and $t_0$, which will be described in III.B. For each trace, we estimate the reference times using Eq.~\ref{eq:LMO}. Then, the gather can be cropped to a smaller sub-gather with the height of 128, in which the center times of each trace are the estimated reference times. 
To ensure the robustness of the input distribution, we apply trace-wise normalization to the data after the LMO correction has been performed. 
Trace-wise normalization is a commonly used technique in seismic data processing\cite{hu2019first}, where the amplitude of each trace are normalized to $[-1, 1]$:
\begin{equation}
    s^{\text{TN}}(x, t) = \frac{s(x, t)}{\max_{1\leq i \leq T}{|s(x, i)|}},
 \label{eq:tw_norm}
\end{equation}
where $s(x, t)$ refers to the amplitude of the seismic wave when the offset to $x$ and the sampling time is $t$, and $T$ represents the total sampling time.
To facilitate the training and testing of our model, we remove all unpicked traces and divide the remaining traces with picks into a gather with a size of 256.
Therefore, the gather after preprocessing is with a size of 128 $\times$ 256.

To detect the signal of FB, we develop a novel U-shape semantic segmentation network based on U-Net, a popular semantic segmentation model in medical image processing\cite{ronneberger2015u}. We develop a dynamic snake convolution (DSConv) module based on DSConv\cite{qi2023dynamic}, and embed it into U-Net. Thus, we call the proposed network to DSU-Net.
As shown in Fig.~\ref{fig:DSUNet}a, DSU-Net includes three down-sampling operations, three up-sampling operation, and three skip-connection operations. 
At the beginning of encoding, low-level features contain a large number of tubular features. Thus, we use two DSConv modules in the first encoder layer. In the second, third, forth encoder layers, we use traditional convolutional module (Fig.~\ref{fig:DSUNet}b) to encode high-level semantic information. The input gather, after three down-sampling operations, becomes one-eighth of the original size, i.e., (16, 32).
In the decoder, we concatenate the encoding features with the decoding features to obtain richer texture information, and the higher-level semantic features are decoded into FB-segmentation map with the same size as the input gather by using three up-sampling operations and one 1$\times$1 Conv layer. We will elaborate on the details of each module below. More details of input and output shapes can be found in Fig.~\ref{fig:DSUNet}.

First, we construct a traditional convolution (TraConv) module to encode or decode the features as shown in Fig.~\ref{fig:DSUNet}b, in which a 3$\times$3 traditional Conv layer, a batch-normalization (BN) layer and an exponential linear unit (ELU) activation function are used in sequence. Exceptionally, since the amplitude of seismic traces is in the range of $[-1, 1]$, we choose the ELU activation function to keep the mapping range of the module output in the same range. Moreover, in each encoding layer, we use two TraConv modules orderly: the first TraConv module compresses the channel and keeps the output size unchanged; The second TraConv module keeps both the number of channels and the size the same.
Second, the DSConv module is designed to extract three parts of features: x-direction texture, y-direction texture, and the local shape. Fig.~\ref{fig:DSUNet}b illustrates that we utilize a x-direction DSConv layer, a y-direction DSConv layer, and a traditional Conv Layer to encode the x-direction texture, y-direction texture, and the local feature, respectively. Concretely, the kernel sizes of these DSConv layers are set to 3$\times$3, aiming to capture long-distance textures. Under the DSConv layer, the corresponding receptive field is 9 in the vertical or horizontal direction. 
Additionally, the kernel size of the traditional Conv layer is 3. Subsequently, we use a BN layer and an ELU activation function to balance the distribution of features and map output features into [-1, 1] nonlinearly, respectively. After encoding three parts of features, we concatenate them and use a TraConv module to fuse the concatenated features, outputting an encoded feature with the same size as the input feature. 
Third, we use a 2$\times$2 transpose Conv layer and an ELU activation function to conduct the up-sampling operation, denoted as Up-Sampling Module as shown in Fig.~\ref{fig:DSUNet}d.plays the seismic traces o
Finally, we utilize a 1$\times$ 1 Conv layer and a Sigmoid activation function to map the decoded features with the size of (32, 128, 256) to the segmentation map with the same of (128, 256) as well as compress the value range into [0, 1] nonlinearly.

During the model training process, the DSU-Net receives a gather as input and produces a binary segmentation map as output. To supervise the output of FB segmentation, there are two methods of generating ground-truth based on manual pickings: On the one hand, pixels representing FB on each trace are marked by 1, while the rest are set to 0. On the other hand, the pixels on each trace from the FB and afterward to 1, while the pixels before the FB are set to 0. For the DSU-Net, since the DSConv detects the continuous changes of FBs among traces, we adopt the former labelling method to supervise the DSU-Net output.

Particularly, to assess the effectiveness of shallow feature extraction, we compared two different strategies: conducting the DSConv module in the first layer of the encoder or using them in both the first and second layers of the encoder. The experimental results indicate that the performance of utilizing the DSConv module solely in the first layer of the encoder exhibits superior generalization capabilities. Moreover, the computational burden during training is notably reduced compared to employing DSConv modules in two layers. Therefore, we adopt the former strategy in our architecture.

The last step of our picking framework is estimating the FB of each trace from the segmentation map of DSU-Net output. Concretely, we select the pixels corresponding to the maximum values of each trace as the FB times. Additionally, we also set a manual threshold to filter the unreasonable picking. When the maximum value of the trace is below the threshold, the picking of this trace will be removed. 

\section{Experiments}
\subsection{Dataset and Metrics}
\begin{figure}[!ht]
    \centering
    \subfloat[Lalor]{\includegraphics[width=0.5\linewidth]{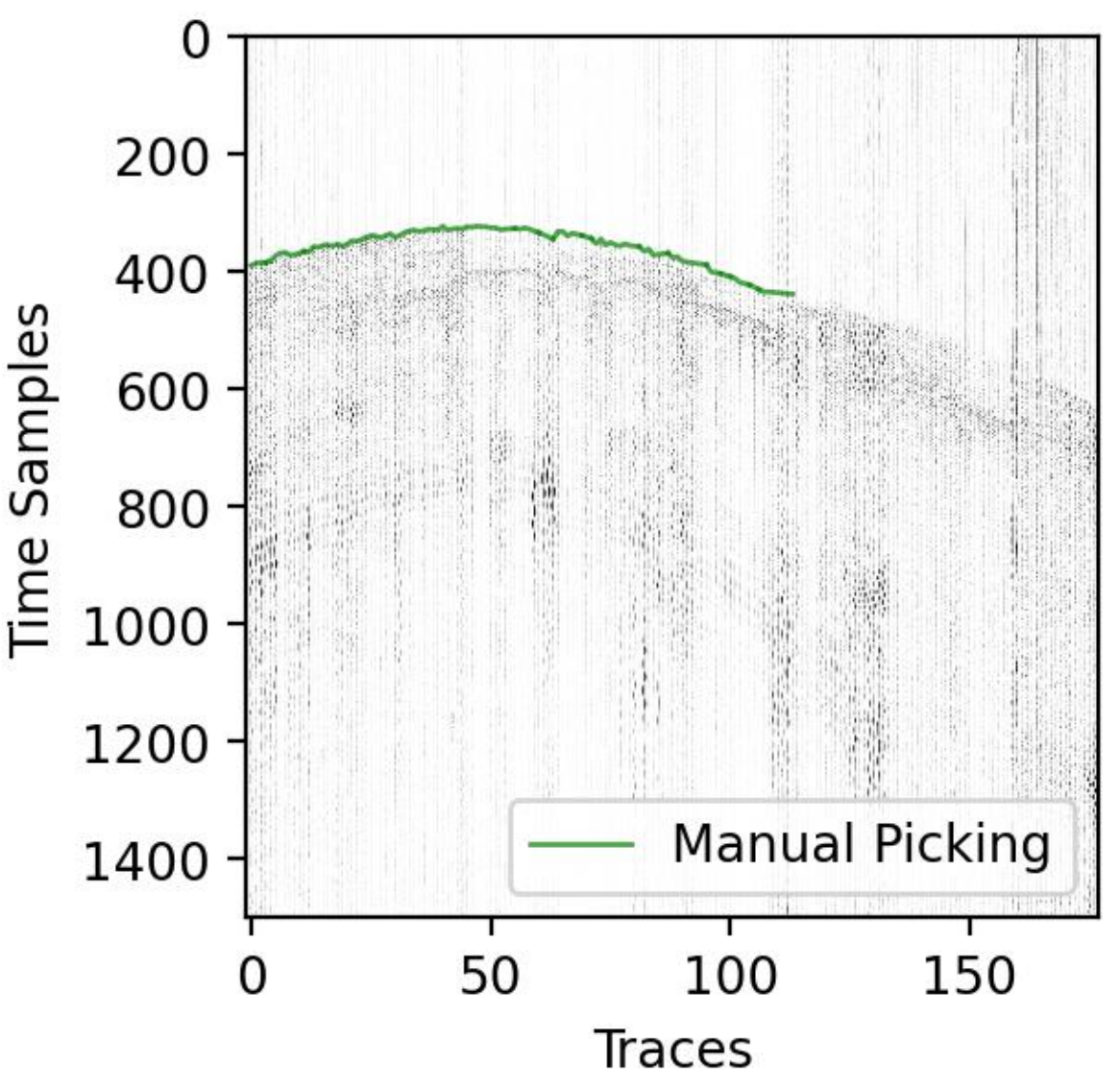}\label{fig:data_L}}
    \subfloat[Sudbury]{\includegraphics[width=0.5\linewidth]{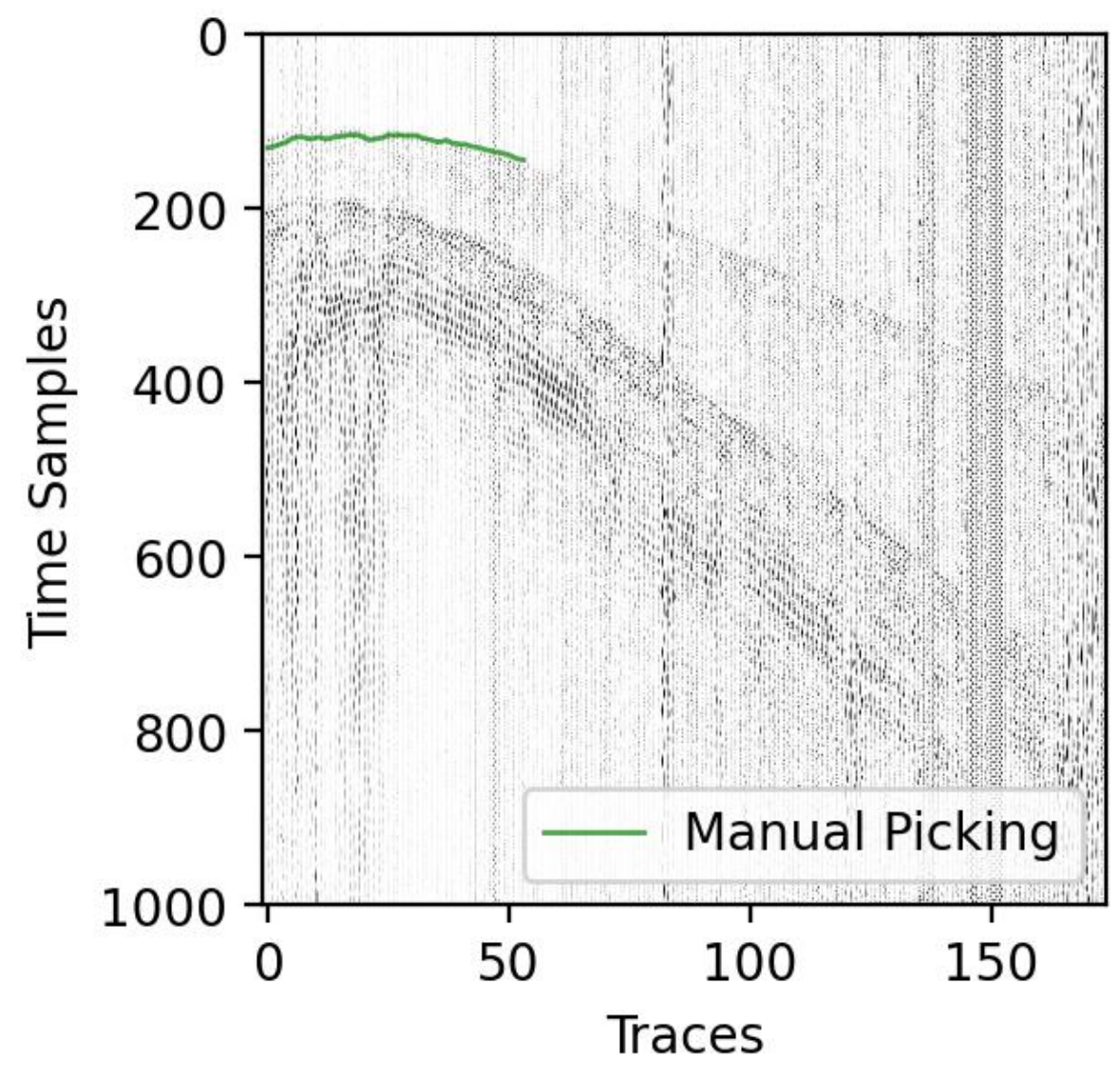}\label{fig:data_S}}\\
    \subfloat[Brunswick]{\includegraphics[width=0.5\linewidth]{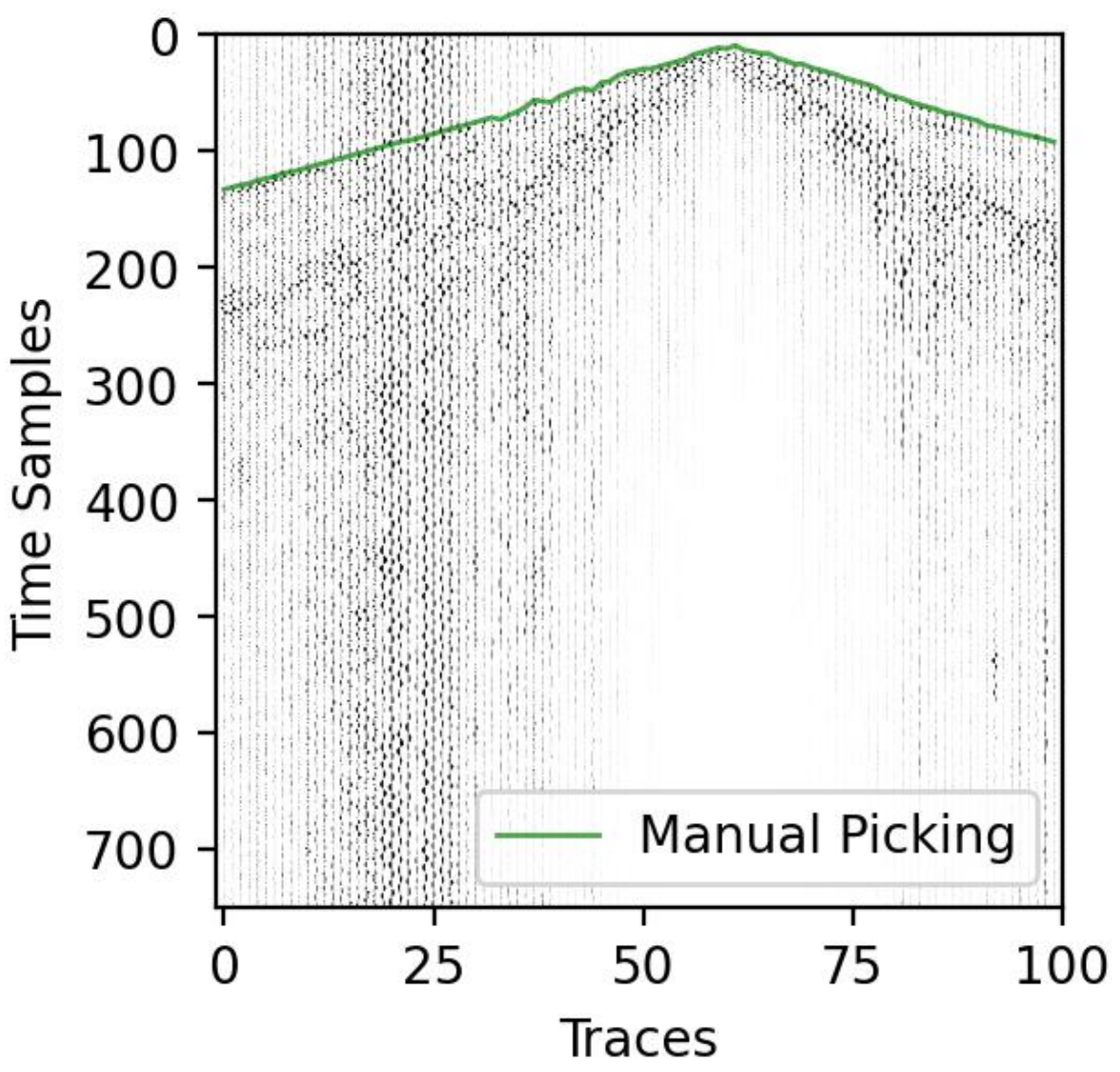}\label{fig:data_B}}
    \subfloat[Halfmile]{\includegraphics[width=0.5\linewidth]{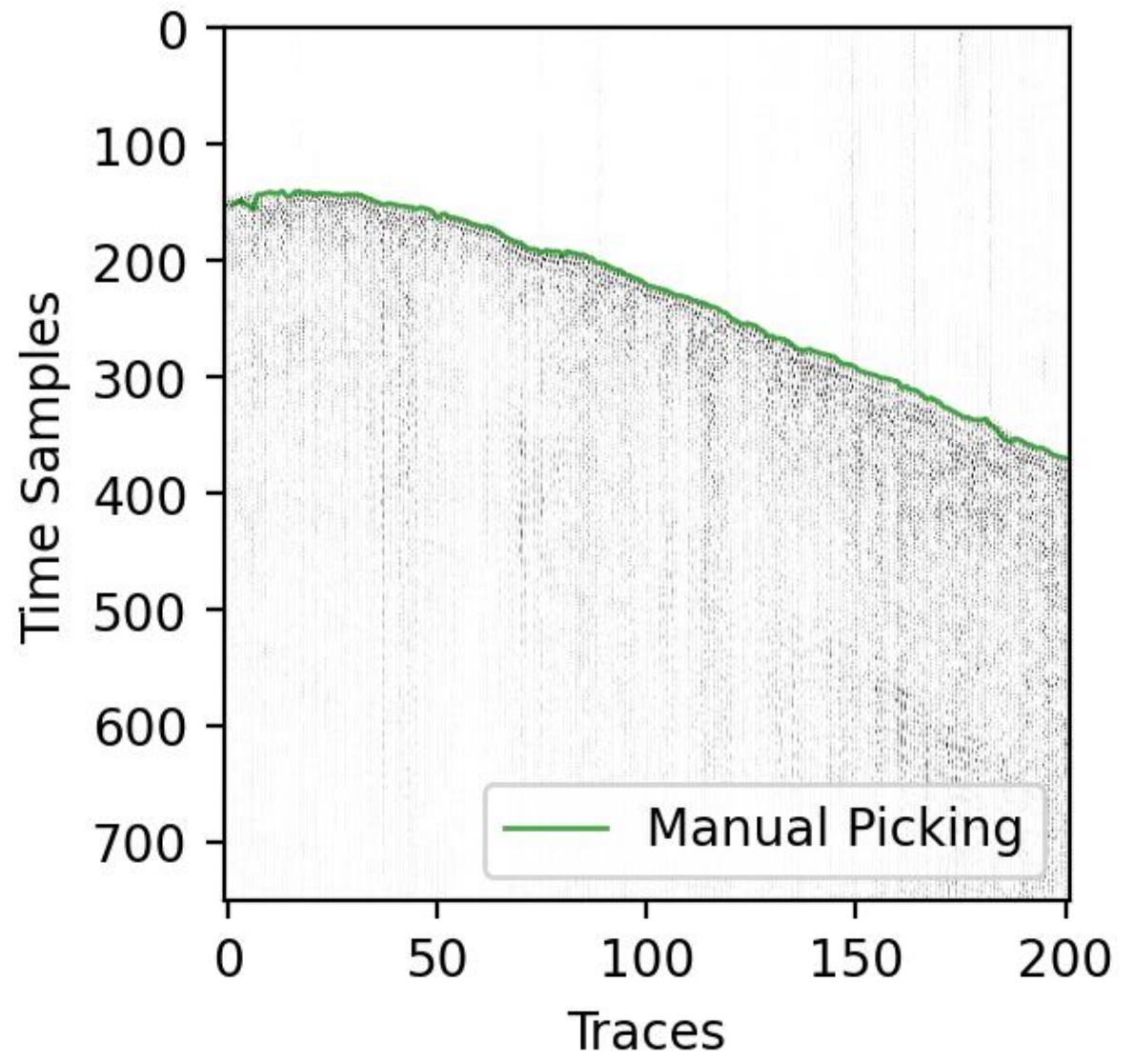}\label{fig:data_H}}
    \caption{Four classic shot gathers of Lalor, Sudbury, Brunswick, and Halfmile. The manual picking is marked on each sub-figure using the green curves.}
    \label{fig:datasets}
\end{figure}

To validate DSU-Net, we choose a popular open-source public data set\cite{st2024deep}, containing four specific field surveys named Halfmile, Lalor, Brunswick, and Sudbury. The experimental results on these datasets can be better compared with previous work and provide a valuable reference for future studies. Specifically, Fig.~\ref{fig:data_L} displays the seismic traces of a shot gather in Lalor. It is evident that the SNR of the traces is low, and there is a significant amount of noise in the far offset region. Consequently, human experts would not pick FB in far-offset regions with low SNR. 
Then, Fig.~\ref{fig:data_S} indicates that most traces have a low SNR in the Sudbury. For instance, the traces in the range of 50-100 exhibit a low SNR, where the FBs are obscured by noise, making them challenging to pick. Therefore, in Sudbury, only a few traces with higher SNR are picked by human experts. Moreover, Fig.~\ref{fig:data_B} and \ref{fig:data_H} show classic seismic shot gathers from Brunswick and Halfmile, respectively, illustrating that both Brunswick and Halfmile have a higher SNR and a higher manual picking ratio. 
Therefore, these four datasets have their characteristics, which can thoroughly verify the performance of the automatic picking model. 

To evaluate the difference between automatic and manual pickings, we utilize the same quantitative metrics as the benchmark model paper \cite{st2024deep}. Besides, we observe variations in the picking rates of the automatic pickers when different segmentation thresholds are applied. Therefore, we first define the automatic picking rate (APR) as the one described in \cite{wang2024msspn} to measure the picking rates:
\begin{equation}
    \text{APR} = \frac{1}{N}\sum\limits_{i=1}^N I\left\{t^A_i>0\right\},
 \label{eq:APR}
\end{equation}
where $t^A_i$ represents $i$-th FB of automatic pickings and $N$ indicates the total number of traces.
Next, we define metrics in \cite{st2024deep} based on the intersection traces of automatic and manual pickings. To evaluate the accuracy of automatic pickings under various error level, $\delta$-dependent hit rate (HR@$\delta$px) is defined by:
\begin{equation}
    \text{HR}@\delta\text{px} = \frac{1}{N}\sum\limits_{i=1}^N I\left\{|t^A_i-t^M_i|<\delta\right\},
 \label{eq:HRkpx}
\end{equation}
where $t^M_i$ represents $i$-th FB of manual pickings and $\delta$ can be set 1, 3, 5, 7, and 9. Particularly, HR@1px represents the accuracy of the absolute error being 0. 
Subsequently, to evaluate the mean absolute error, relative error, and stability, the mean absolute error (MAE), mean bias error (MBE), and root mean square error (RMSE) are defined, respectively:
\begin{equation}
    \text{MAE} = \frac{1}{N}\sum\limits_{i=1}^N |t^A_i-t^M_i|,
 \label{eq:MAE}
\end{equation}
\begin{equation}
    \text{MBE} = \frac{1}{N}\sum\limits_{i=1}^N \left(t^A_i-t^M_i\right),
 \label{eq:MBE}
\end{equation}
\begin{equation}
    \text{RMSE} = \left(\frac{1}{N}\sum\limits_{i=1}^N \left(t^A_i-t^M_i\right)^2\right)^{1/2}.
 \label{eq:RMSE}
\end{equation}

\subsection{Implementation Details}
This section covers four aspects: dataset settings, training hyperparameter settings, LMO hyperparameter settings, and a novel data augmentation method.
First, our study utilizes the same four-fold experiment as \cite{st2024deep} to compare with the benchmark method, and the four-fold setting of the training sets, the validation set and the test set are illustrated in Tab.~\ref{tab:datasets}.
\begin{table}[!ht]
\centering
\caption{Open dataset information}\label{tab:datasets}
\resizebox{0.4\textwidth}{!}{\begin{tabular}{lllll}
\toprule 
Fold & \multicolumn{2}{c}{Train Sets} & Validation Set & Test Set  \\ \hline
\#1   & Halfmile       & Lalor         & Brunswick      & Sudbury   \\
\#2   & Sudbury        & Halfmile      & Lalor          & Brunswick \\
\#3   & Lalor          & Brunswick     & Sudbury        & Halfmile  \\
\#4   & Brunswick      & Sudbury       & Halfmile       & Lalor    \\ \bottomrule
\end{tabular}}
\end{table}

Second, the primary hyperparameters in the training process include the epoch number, the optimizer, the learning rate, the weight decay, and the loss function. The candidate values are shown in Tab.~\ref{tab:train_para_tuning}. Specifically, we perform the model training five times with different random initial seeds on each fold. Based on the model performance on the validation set, we determine relatively optimal training hyperparameters marked in bold in Tab.~\ref{tab:train_para_tuning}.
\begin{table}[!ht]
\centering
\caption{Tuning Hyperparameters of Training Process}\label{tab:train_para_tuning}
\resizebox{0.4\textwidth}{!}{\begin{tabular}{lc}
\toprule 
Hyperparameters & Possible Value                     \\ \hline
Epoch Number    & 10,15,\textbf{20}                           \\
patience        & 2,\textbf{4},6                              \\
Decoder Blocks  & $\left[512,256,128,64\right]$, \boldmath $\left[256,128,64,32\right]$ \\
Optimizer       & \textbf{Adam},AdamW,SGD                     \\
Learning rate   & 0.1,0.01,\textbf{0.001}                     \\
Weight decay    & $10^{-5}$, \boldmath $10^{-6}$                    \\
Loss function   & \textbf{Binary Cross-Entropy}, Dice          \\ \bottomrule
\end{tabular}}
\end{table}

Third, the LMO correction requires determining two hyperparameters: the reference velocity $v$ and the interception time $t_0$. Our study uses very few manual pickings to estimate $v$ and $t_0$, and Tab.~\ref{tab:LMO_setting} indicates the specific setting. 
\begin{table}[!ht]
\centering
\caption{Tuning Hyperparameters of Training Process}\label{tab:LMO_setting}
\resizebox{0.4\textwidth}{!}{\begin{tabular}{lcc}
\toprule 
Dataset   & Reference Velocity (m/s) & Interpolation Time (s) \\\hline
Lalor     & 6013.97                  & 0.0130                 \\
Halfmile  & 5349.20                  & 0.0234                 \\
Brunswick & 5136.20                  & 0.0017                 \\
Sudbury   & 5891.69                  & 0.0367                 \\\bottomrule
\end{tabular}}
\end{table}

Finally, we propose a novel data augmentation method based on LMO correction. To enhance the robustness and generalization ability of the model, we augment the input gathers in the training sets during the training process. Specifically, based on Eq.~\ref{eq:LMO}, we initially generate the corresponding normal distributions using the values of $t_0$ and $v$, along with their standard deviations estimated by the manual picking of a few shot gathers. Subsequently, we randomly select $t_0^{\text{aug}}$  and $v^{\text{aug}}$ from the normal distribution, substitute Eq.~\ref{eq:LMO}, and add the random perturbation $s^{\text{rand}}$ to get a new cropping value $t^{\text{aug}}$.
\begin{equation}
    t^{\text{aug}} = t_0^{\text{aug}} + \frac{x}{v^\text{aug}} + s^{\text{rand}} \cdot \Delta t
 \label{eq:data_aug}
\end{equation}
where $x$ is the offset of the augmented trace, $\Delta t$ is the sampling interval of the time axis, and $s^{\text{rand}}$ is sampled from a uniform distribution between -32 and 32.

\subsection{Comparative Experiments}
\subsubsection{Introduction of Comparative Methods}
Our study compares the proposed DSU-Net with the classical STA/LTA model, the U-Net-based benchmark method\cite{st2024deep}, and the STU-Net\cite{jiang2023seismic}. The STA/LTA method is a deterministic model used to detect the FB by calculating whether the STA/LTA value surpasses a predefined threshold. 
The benchmark method consists of two parts: an encoder and a decoder. The encoder extracts features from the input image, while the decoder reconstructs the segmentation results of the image. 
The STU-Net also adopts an encoder-decoder architecture connected by a dilated convolution block. The encoder part of STU-Net utilizes a variant convolution module
in Swin Transformer called the Swin-T block\cite{liu2021swin} to extract hierarchical features from the input image. The SwinT block extends the traditional one-dimensional sequence of the Transformer into two-dimensional image patches. Then, it employs a hierarchical attention mechanism to capture features within a larger receptive field. Additionally, the SwinT block introduces a window shifting mechanism on top of the self-attention mechanism, limiting the attention computation to windows in the vicinity of the current region preserves positional information better. 

For the STA/LTA method, we select the long and short window lengths from [50, 40, 30, 20, 10] and [5, 4, 3, 2], respectively. After test experiments, the optimal parameters are determined based on the MAE metrics: a long window length of 30 and a short window length of 3. 
Additionally, the model hyperparameters and training hyperparameters of the benchmark method and the STU-Net are consistent with the proposed setting in the original paper\cite{st2024deep}\cite{jiang2023seismic}. 
For each fold, the benchmark method, STU-Net, and DSU-Net are trained using the training hyperparameters with ten various seeds. Subsequently, we test each training model of each method respectively, and calculate the mean and standard variance of the ten test results as the final quantitative results. However, the picking thresholds of the above mentioned models determine the APRs and other metrics of automatic picking. Generally, when the APR goes down, the HR@$\delta$px goes up.
Thus, we need to compare other metrics with the same APR to compare models fairly.

\begin{table*}[!ht]
\centering
\caption{Four-Fold Experimental Results of Comparative Models}\label{tab:q_comp}
\resizebox{0.9\textwidth}{!}{\begin{tabular}{llcccccccc}
\toprule
Test Site & Method          & HR@1px   & HR@3px   & HR@5px   & HR@7px   & HR@9px   & RMSE    & MAE     & MBE     \\\hline
Sudbury   & STA/LTA         & 28.8    & 52.9    & 81.9    & 87.3    & 88.4     & 10.8    & 5.1     & -0.2    \\
          & U-Net benchmark\cite{st2024deep} & 94.9±0.4 & 95.3±0.2 & 95.9±0.1 & 96.7±0.1 & 97.1±0.1 & 5.8±0.1 & 1.0±0.0     & 0.2±0.0   \\
          & STU-Net\cite{jiang2023seismic}         & 94.4±0.3 & 95.2±0.1 & 95.8±0.0   & 96.6±0.0   & 97.1±0.0   & 5.9±0.0   & 1.0±0.0     & 0.2±0.0   \\
          & DSU-Net         & \textbf{95.2±0.1} & \textbf{95.5±0.1} & \textbf{95.9±0.1} & \textbf{96.7±0.0}   & \textbf{97.1±0.0}   & \textbf{5.9±0.0}   & \textbf{1.0±0.0}     & 0.2±0.0   \\
Brunswick & STA/LTA         & 38.2    & 56.8    & 86.5     & 96.7    & 97.4    & 4.6     & 2.5     & 1.6     \\
          & U-Net benchmark\cite{st2024deep} & 97.1±1.6 & 99.6±0.1 & 99.8±0.1 & 99.9±0.0   & 99.9±0.0   & 0.5±0.1 & 0.1±0.0   & 0.0±0.0     \\
          & STU-Net\cite{jiang2023seismic}         & 95.3±3.8 & 99.3±0.1 & 99.6±0.1 & 99.8±0.0   & 99.9±0.0   & 0.6±0.1 & 0.1±0.0   & 0.0±0.0     \\
          & DSU-Net         & \textbf{99.2±0.4} & \textbf{99.7±0.0}   & \textbf{99.9±0.0}   & \textbf{99.9±0.0}   & \textbf{100.0±0.0}    & \textbf{0.4±0.0.0}   & \textbf{0.0±0.0}     & 0.0±0.0    \\
Halfmile  & STA/LTA         & 35.8    & 51.8    & 70.4    & 93.4    & 95.9    & 5.1     & 3.0       & 2.0       \\
          & U-Net benchmark\cite{st2024deep} & 95.4±0.6 & 96.1±0.2 & 97.2±0.1 & 98.9±0.0   & 99.6±0.0   & 1.2±0.0   & 0.2±0.0   & 0.0±0.0     \\
          & STU-Net\cite{jiang2023seismic}         & 94.5±0.8 & 96.1±0.1 & 97.2±0.0   & 98.8±0.0   & 99.6±0.0   & 1.2±0.0   & 0.2±0.0   & 0.0±0.0     \\
          & DSU-Net         & \textbf{96±0.5}   & \textbf{96.4±0.1} & \textbf{97.3±0.1} & \textbf{98.9±0}   & \textbf{99.6±0.0}   & \textbf{1.2±0.0}   & \textbf{0.2±0.0}   & 0.0±0.0     \\
Lalor     & STA/LTA         & 49.0    & 75.5    & 82.7    & 89.4    & 93.0       & 4.5     & 2.1     & 1.7     \\
          & U-Net benchmark\cite{st2024deep} & 83.2±3.5 & 89.1±1.6 & 91.8±1.1 & 95.4±0.8 & 97.5±0.5 & 2.7±0.4 & 0.8±0.2 & 0.5±0.1 \\
          & STU-Net\cite{jiang2023seismic}         & 80.7±4.2 & 87.9±2.1 & 90.9±1.4 & 94.6±1.0   & 97±0.7   & 2.9±0.4 & 0.9±0.2 & 0.5±0.1 \\
          & DSU-Net         & \textbf{86.8±2.1} & \textbf{90.7±1.4} & \textbf{92.8±1.0}   & \textbf{95.9±0.6} & \textbf{97.8±0.4} & \textbf{2.4±0.3} & \textbf{0.7±0.1} & 0.5±0.1 \\\bottomrule
\end{tabular}}
\end{table*}

\begin{figure*}[!ht]
    \centering
    \subfloat[]{\includegraphics[width=0.27\textwidth]{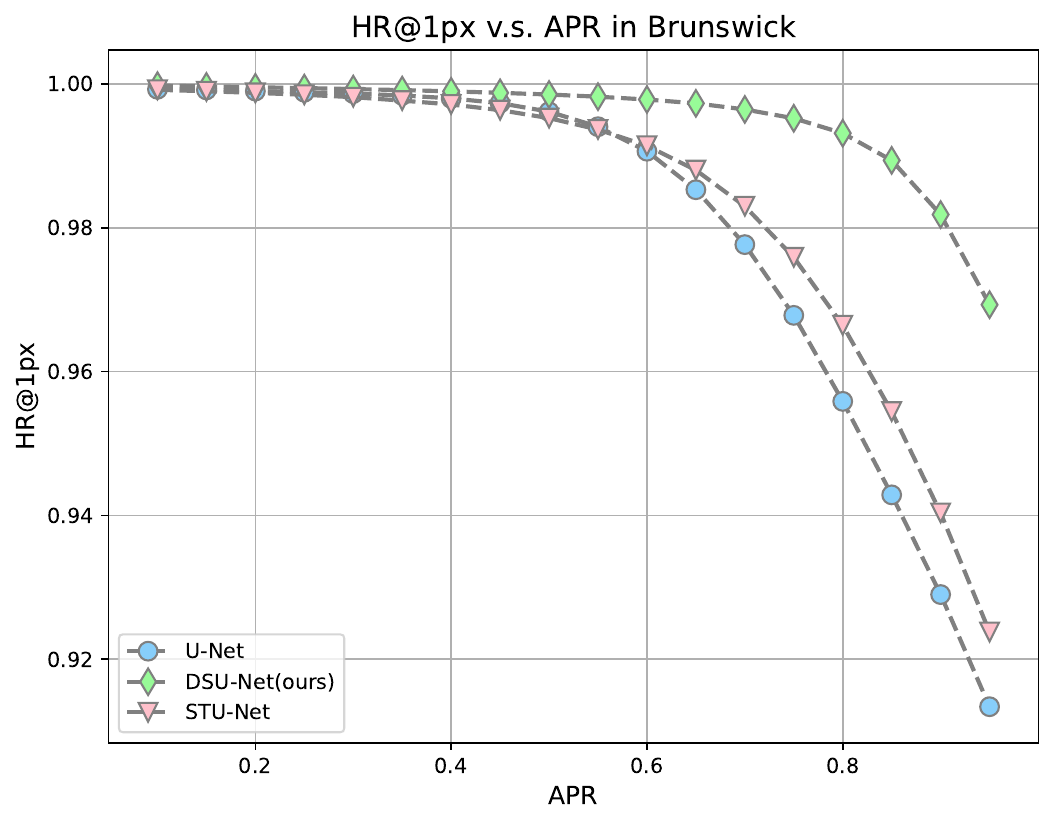}\label{fig:332_HR1}}
    \subfloat[]{\includegraphics[width=0.27\textwidth]{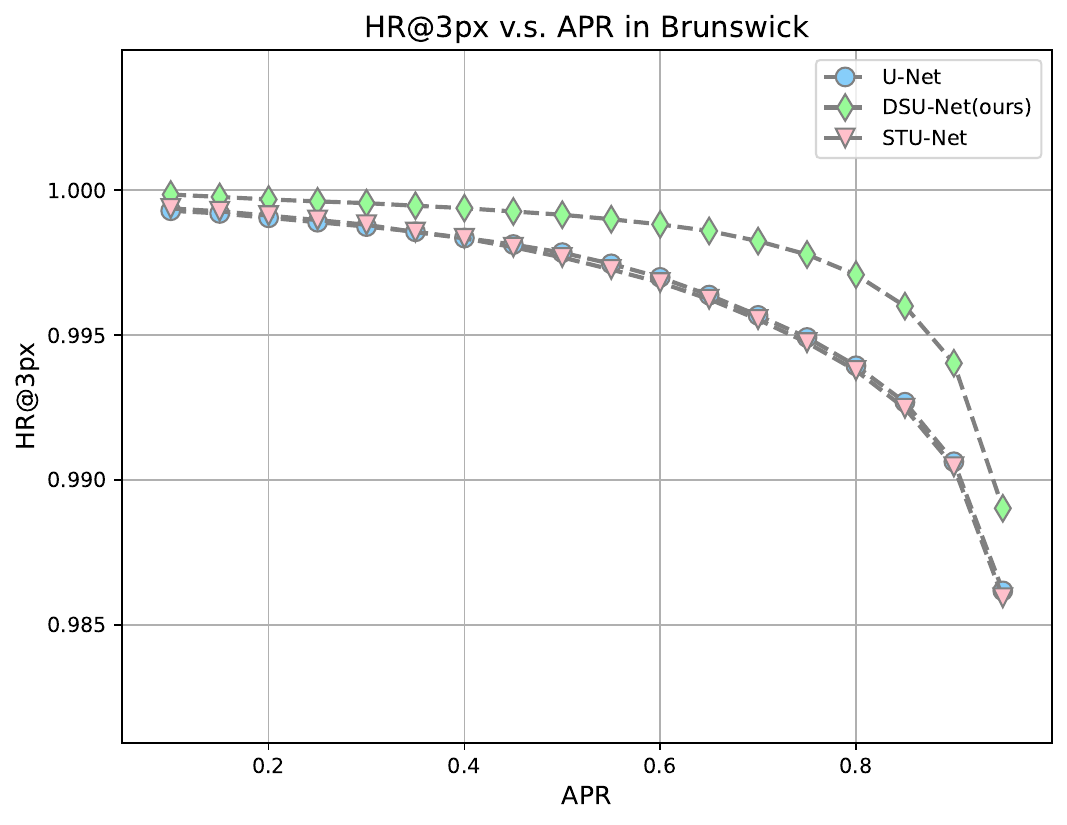}\label{fig:332_HR3}}
    \subfloat[]{\includegraphics[width=0.27\textwidth]{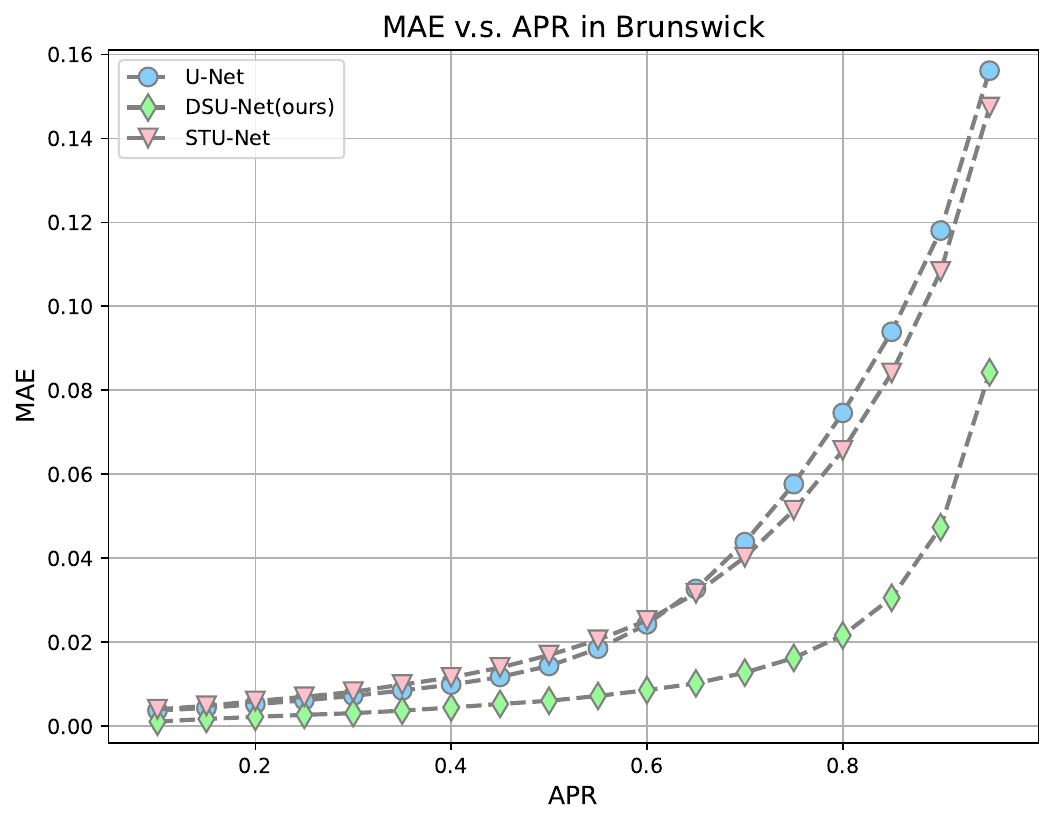}\label{fig:332_MAE}}
    \caption{Comparison of quantitative results of deep learning-based picking methods on Fold \#2.}
    \label{fig:q_comp_v}
\end{figure*}

\subsubsection{Quantitative Comparison}
Tab.~\ref{tab:q_comp} describes the test results with APR = 0.8 of each method under each fold, where the value represents the average value ± standard deviation of ten test results at the same APR (80\%).
It indicates that our method outperforms the other three automatic picking methods on the four test sites. 
Specifically, this can be characterized in terms of picking accuracy and robustness. The Brunswick and Halfmile datasets have higher SNRs and a higher proportion of manual FB annotation, making identifying and picking FB relatively easy. On these two datasets, DSU-Net outperforms other automatic picking methods in terms of picking accuracy, especially on Brunswick, where the HR@1px is over 99\%, far exceeding other methods. This indicates that the DSConv effectively handles fine tubular topological structures, enabling it to better capture subtle features in seismic wave images. Additionally, our approach yields relatively smaller RMSE compared to other methods, illustrating the outstanding picking stability of DSU-Net.
In the test of Lalor (Fold \#4), identifying FB is challenging due to the significant noise presented in the far-offset regions. As shown in Tab.~\ref{tab:q_comp}, DSU-Net offers more stable picking than other methods on Lalor in term of HR@1px and RMSE, demonstrating that under the influence of background noise, it can capture the imperceptible
features of the signal better. Similarly, in the test of Sudbury, where most regions also have a lower SNR, DSU-Net outperforms other models. This highlights the adaptability of DSU-Net across various geological conditions and its exceptional generalization capabilities.

To compare model performance under different APRs more comprehensively, we visualize the relationship between the various APRs and the corresponding measured metrics for the DL-based picking models, as shown in Fig.\ref{fig:q_comp_v}.
The results indicate that, at the same APR, DSU-Net outperforms the other models in terms of HR@1px and HR@3px, particularly for HR@1px. When the APR is high (APR$>$0.8), DSU-Net significantly outperforms the other two models. This suggests that under the severe criterion of the prediction, e.g., being within one pixel of the true result, DSU-Net has a higher accuracy in picking. Fig.~\ref{fig:q_comp_v} also shows that as APR increases, the decline of HR@1px is slower than other models, indicating that our model has better robustness. Moreover, DSU-Net has a lower MAE than the other two models, indicating a smaller average deviation between the automatic picking of DSU-Net and the manual labels. This further confirms the higher picking accuracy of DSU-Net compared to the other models.

In summary, the performance of DSU-Net at all test sites is outstanding, demonstrating its superiority as an effective tool for seismic FB picking. Its unique network design enables it to fully utilize the geometric features of the seismic shot gather. By incorporating the DSConv to handle intricate tubular topological structures, it can extract delicate, steep features and exhibit significant fluctuations. Thus, we conclude that DSU-Net significantly enhances the picking accuracy and holds promising application prospects.

\subsubsection{Visualization Comparison}
\begin{figure}[!ht]
    \centering
    \subfloat[]{\includegraphics[width=0.24\textwidth]{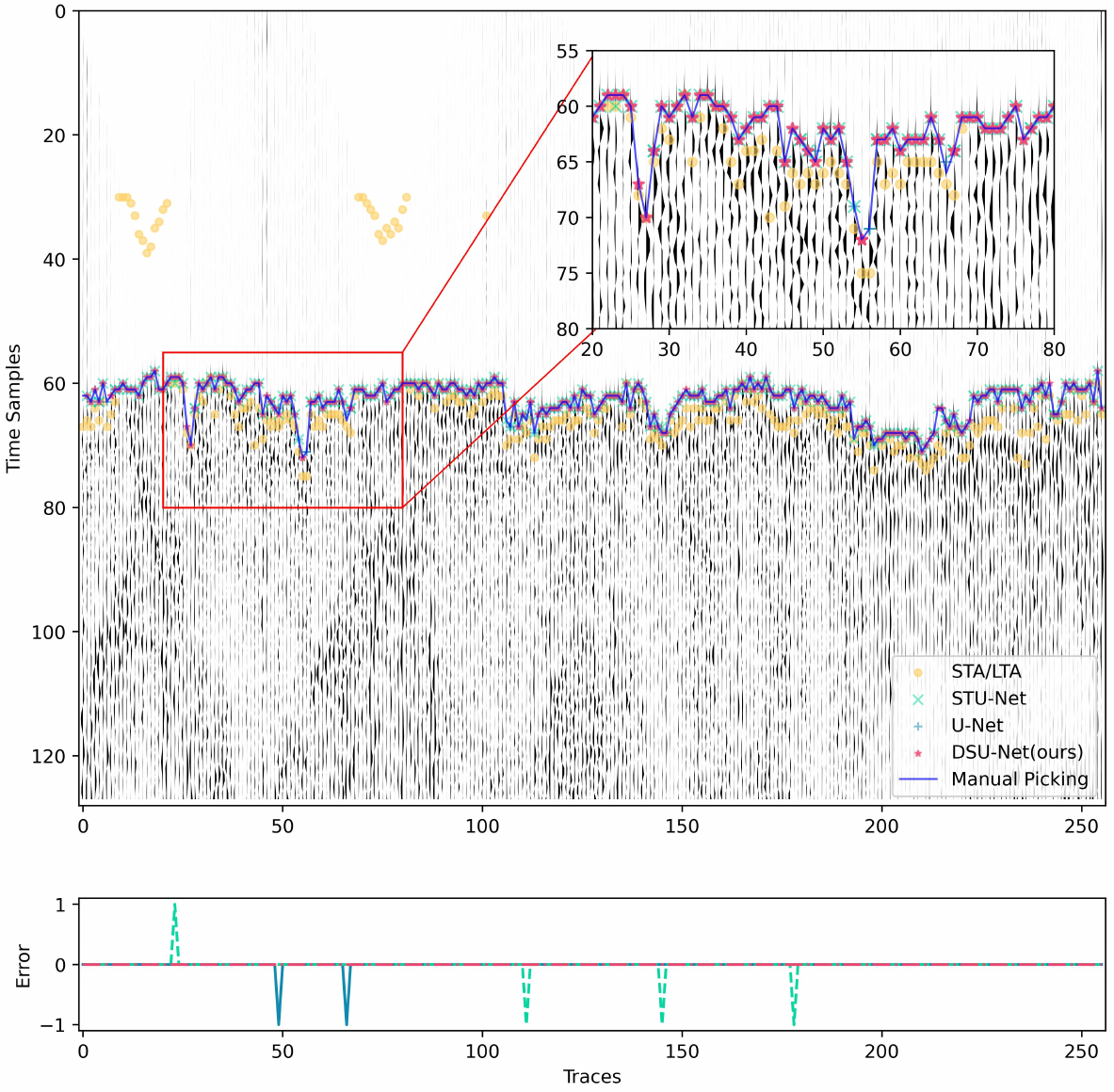}\label{fig:333_S_1}}
    \subfloat[]{\includegraphics[width=0.24\textwidth]{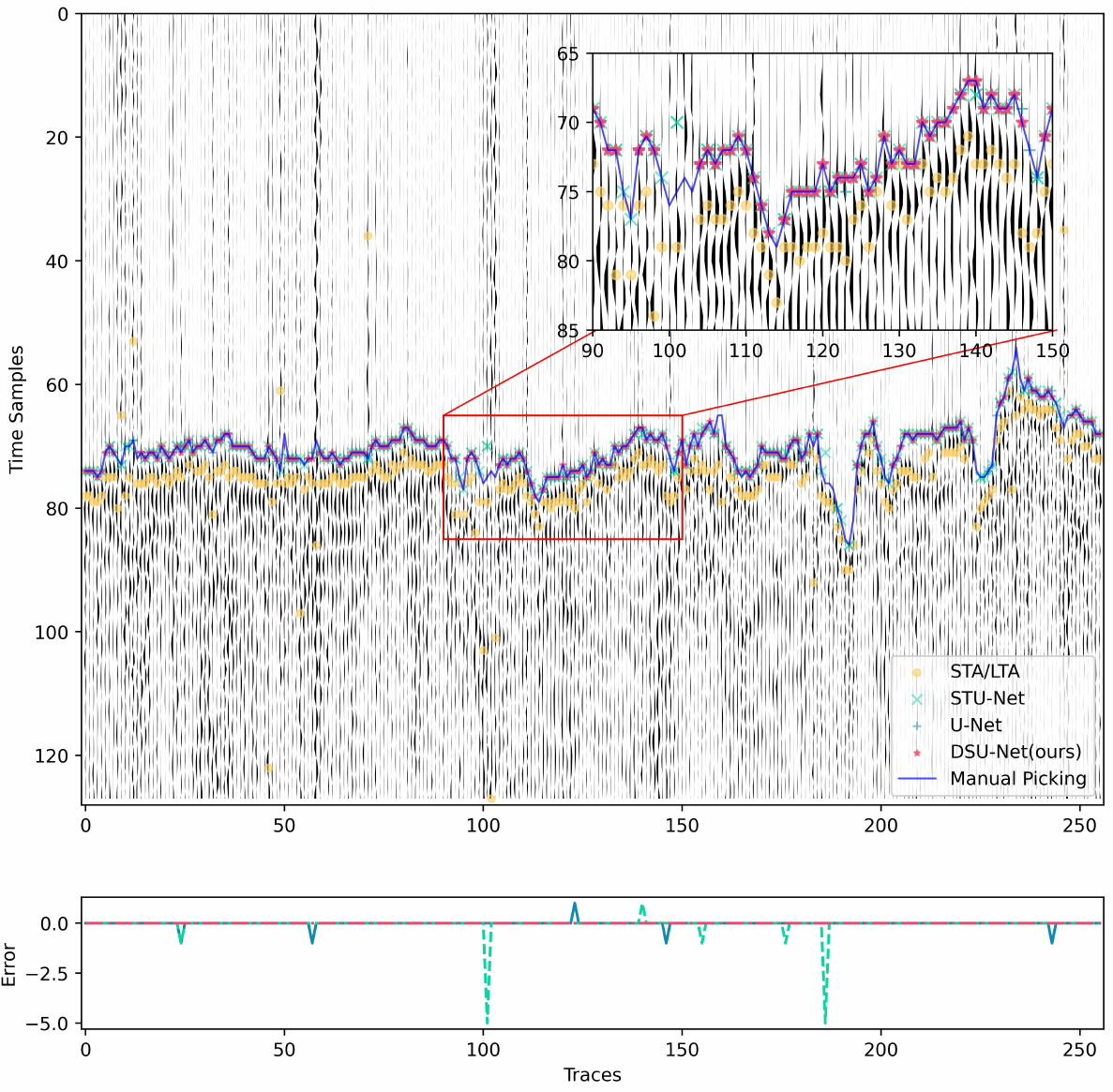}\label{fig:333_S_2}}
    \caption{Picking results of four models and manual picking on Sudbury in Fold \#1. }
    \label{fig:vis_Sudbury}
\end{figure}

\begin{figure}[!ht]
    \centering
    \subfloat[]{\includegraphics[width=0.24\textwidth]{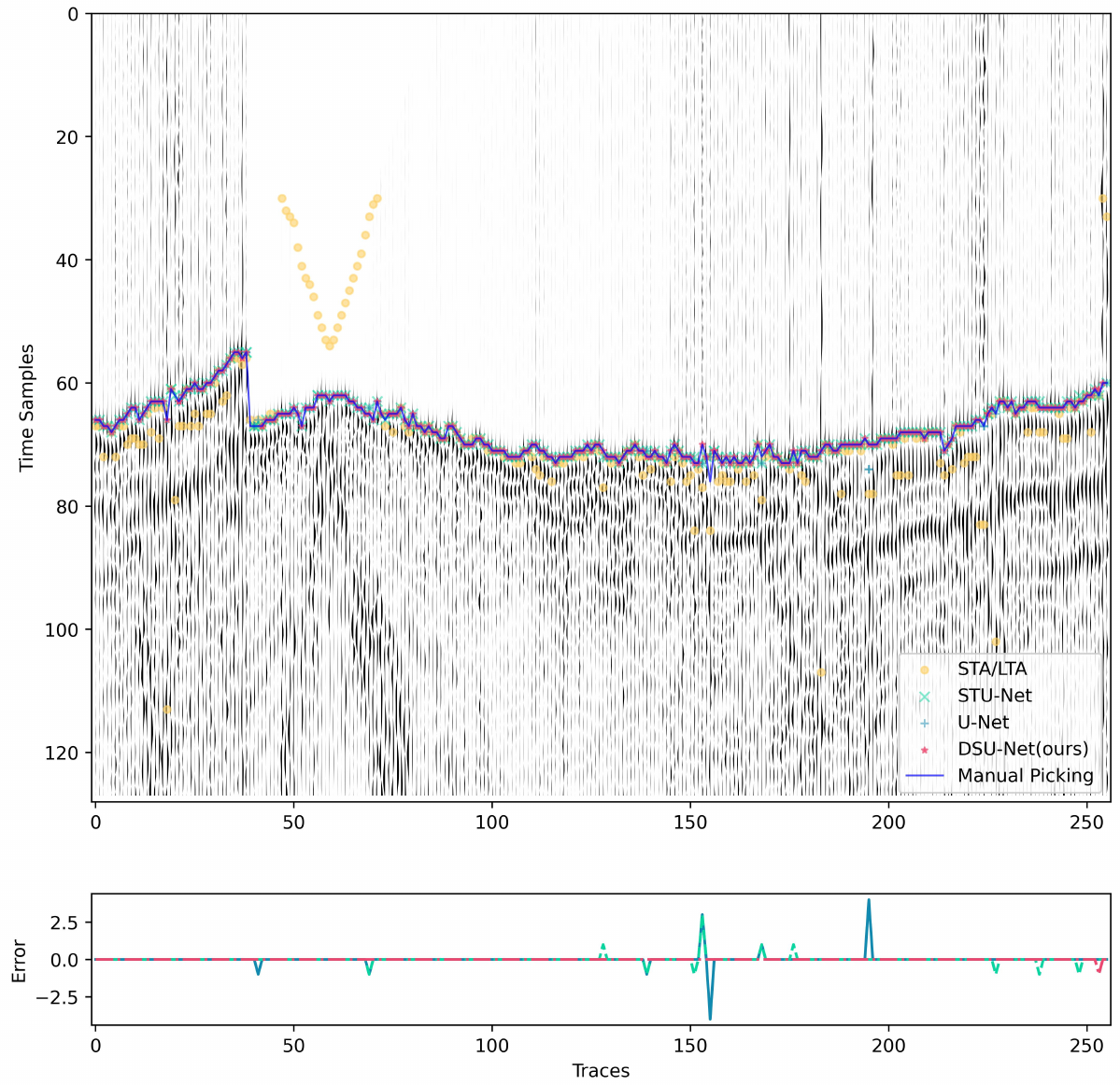}\label{fig:333_B_1}}
    \subfloat[]{\includegraphics[width=0.24\textwidth]{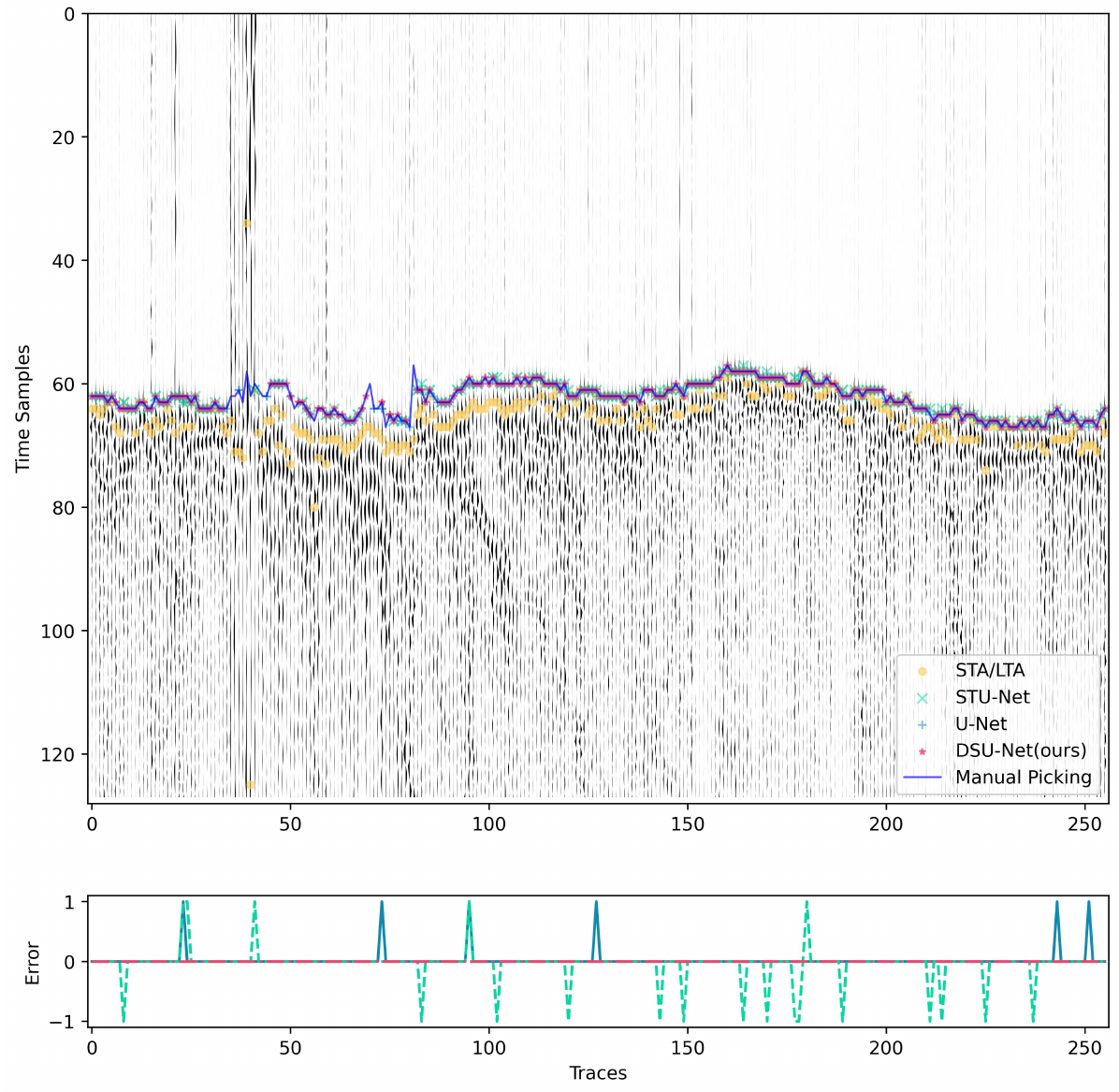}\label{fig:333_B_2}}
    \caption{Picking results of four models and manual picking on Brunswick in Fold \#2. }
    \label{fig:vis_Brunswick}
\end{figure}

\begin{figure}[!ht]
    \centering
    \subfloat[]{\includegraphics[width=0.24\textwidth]{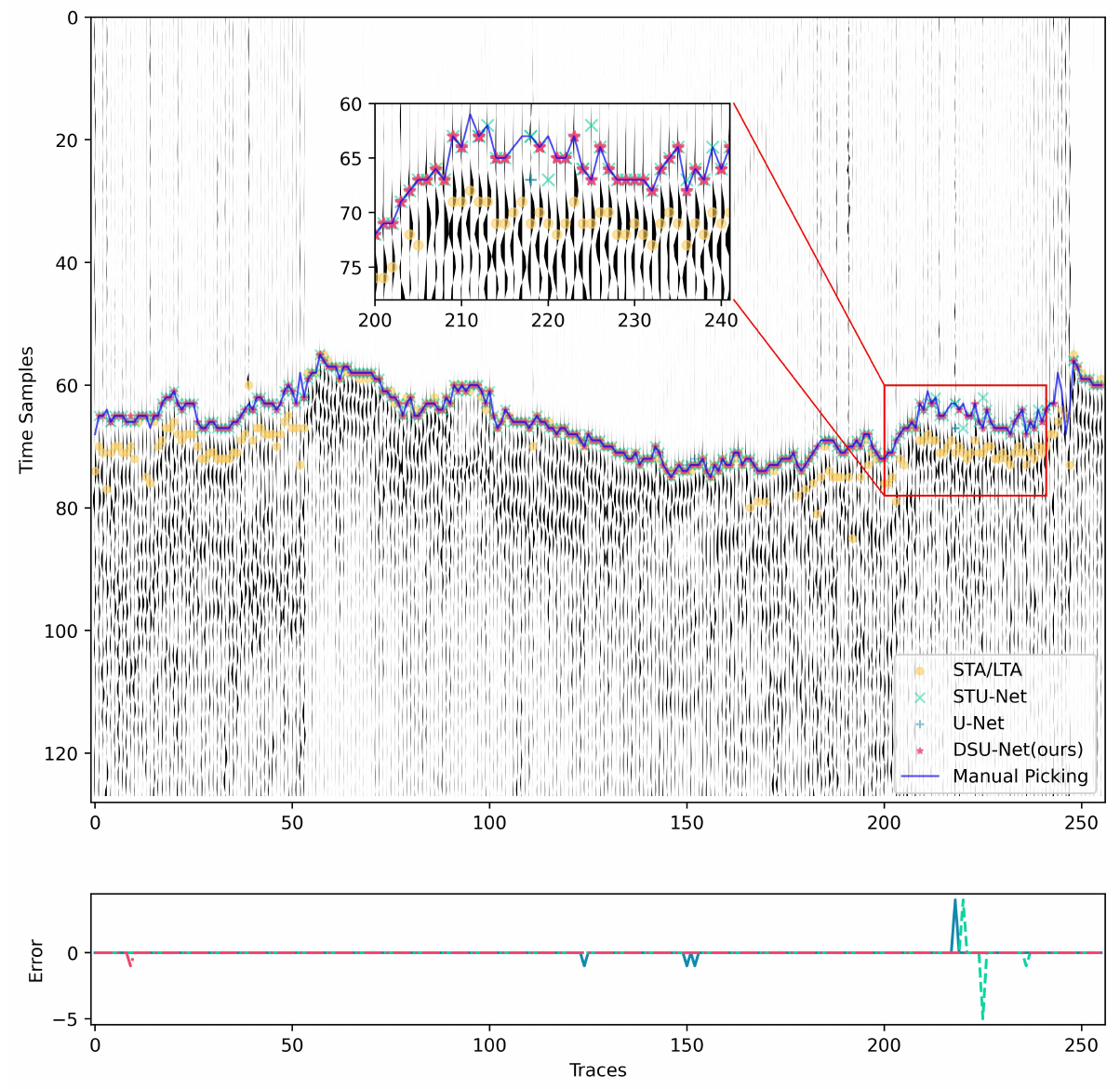}\label{fig:333_H_1}}
    \subfloat[]{\includegraphics[width=0.24\textwidth]{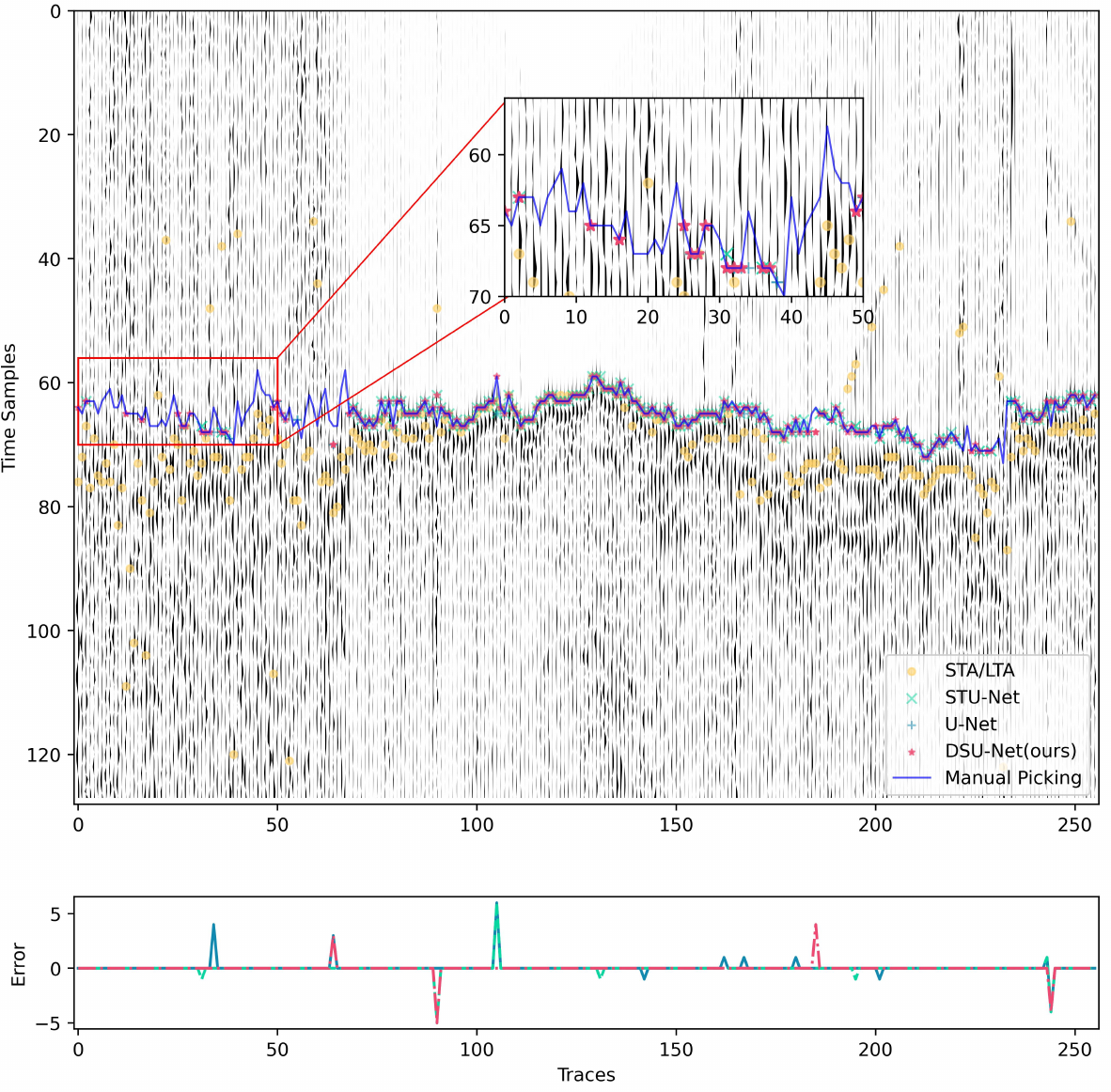}\label{fig:333_H_2}}
    \caption{Picking results of four models and manual picking on Halfmile in Fold \#3. }
    \label{fig:vis_Halfmile}
\end{figure}

\begin{figure}[!ht]
    \centering
    \subfloat[]{\includegraphics[width=0.24\textwidth]{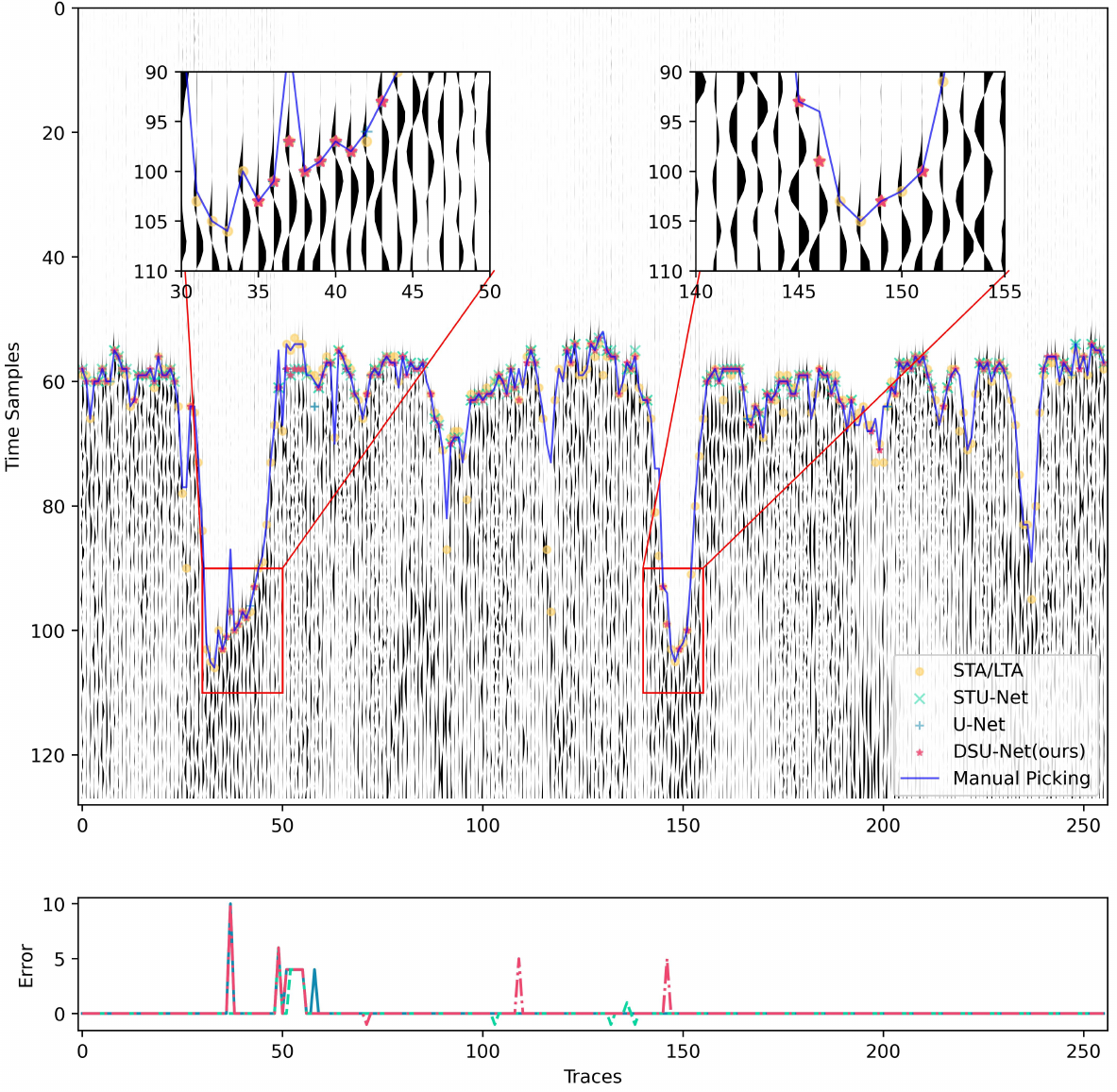}\label{fig:333_L_1}}
    \subfloat[]{\includegraphics[width=0.24\textwidth]{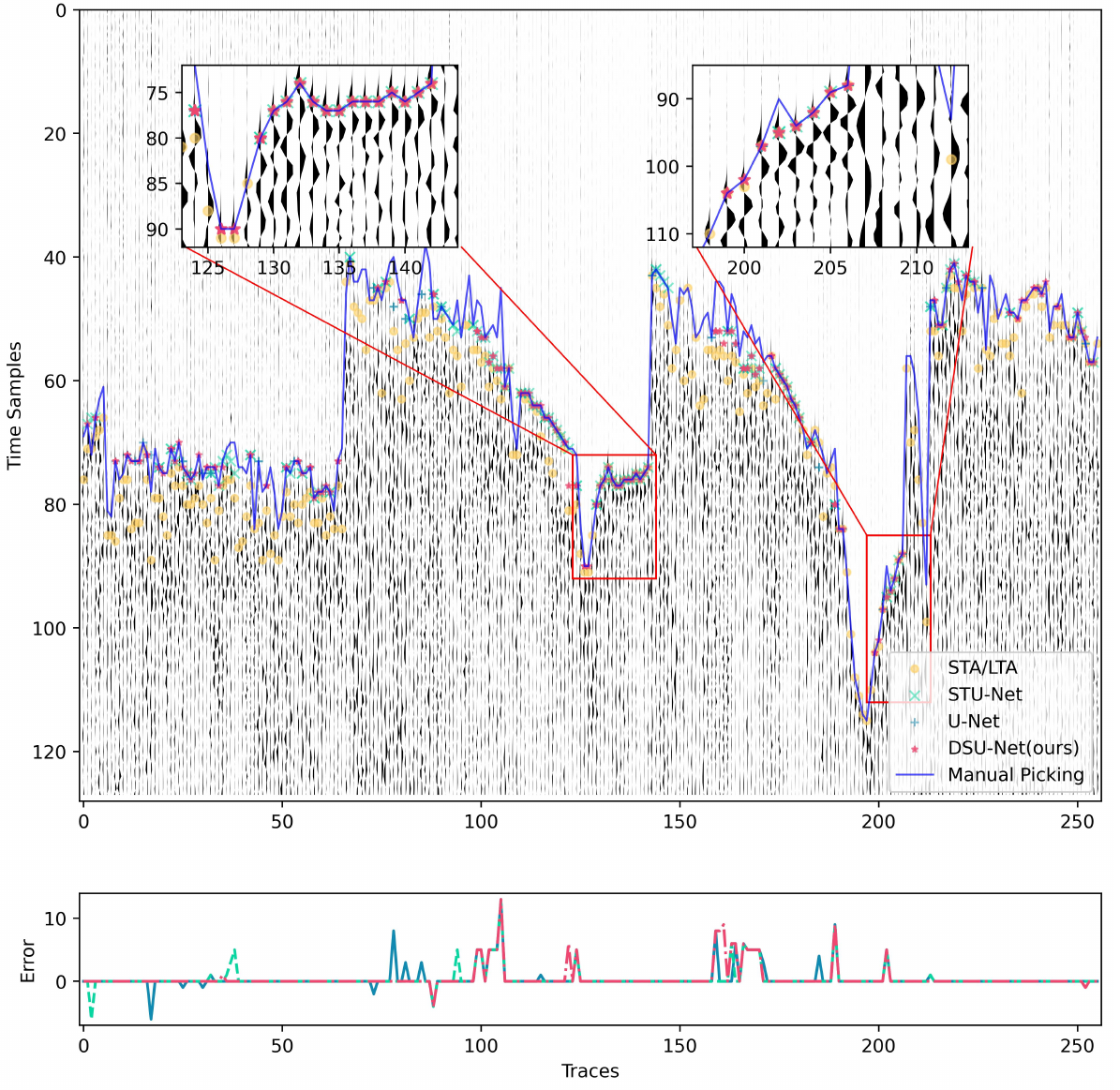}\label{fig:333_L_2}}
    \caption{Picking results of four models and manual picking on Lalor in Fold \#4. }
    \label{fig:vis_Lalor}
\end{figure}

Next, we visualize a few cases on each test site, as depicted in Fig.~\ref{fig:vis_Sudbury}-\ref{fig:vis_Lalor}. DSU-Net achieves relatively robust pickings in these eight examples, even in the Lalor dataset with low SNR.

First, Fig.~\ref{fig:vis_Sudbury} visualizes the FB picking of each model in Sudbury. Upon examining Fig.~\ref{fig:333_S_1}, it is evident that DSU-Net can provide stable and accurate FB pickings under conditions of a high SNR and minimal changes in the shot gather. Even when the shot gather undergoes slight changes, DSU-Net can effectively detect these changes promptly. For instance, between traces 20-80, DSU-Net identifies the tubular geometric texture, demonstrating its ability to respond to abrupt changes in the feature map along the y-axis direction from a previously stable and continuous state. At the same time, we can conclude that the picking results of DSU-Net in Sudbury are more robust based on the error analysis chart (the bottom sub-figure in Fig.~\ref{fig:333_S_1}). Moreover, at the 90-150 trace in Fig.~\ref{fig:333_S_2}, both U-Net and STU-Net are influenced by background noise to different extents, affecting the accuracy of FB. In contrast, DSU-Net exhibits an anti-noise characteristic.

Second, Fig.~\ref{fig:vis_Brunswick} displays the results of FB pickings by various models at the Brunswick site. The Brunswick dataset has a relatively high SNR, which gives us reason to assume that the manual FB pickings are accurate enough. Additionally, various models achieve relatively high APR on this dataset. Therefore, we measure the accuracy of the model predictions by comparing the relative error between the automatic FB and the manual FB. The error analysis graph shows that the DSU-Net has a smaller relative error than the other models, demonstrating the precision of the DSU-Net's pickings. The figure also illustrates that when the shot gathers are relatively stable along the x-axis, the DSU-Net exhibits better stability. 

Third, Fig.~\ref{fig:vis_Halfmile} shows the pickings in Halfmile. As seen in Fig.~\ref{fig:333_H_1}, the accuracy of DSU-Net in FB picking is generally high. In regions with a low SNR (traces in the range of 200-250), DSU-Net accurately detects these traces. In contrast, U-Net barely detects any traces, and STU-Net also shows a poor result. Fig.~\ref{fig:333_H_2} illustrates that in the high-noise region (trace range from 0-50), U-Net and STU-Net detect only a small number of traces, while DSU-Net identifies more traces with high accuracy. It can be observed that DSU-Net exhibits superior stability and better signal acquisition capabilities compared to U-Net and STU-Net in high-noise environments. Additionally, DSU-Net excels in extracting subtle features between the signal and noise.

Finally, Fig.~\ref{fig:vis_Lalor} visualizes the results of each model at the Lalor site. The FB curve changes drastically from traces 30-50, 140-160 in Fig.~\ref{fig:333_L_1}, and traces 195-215 in Fig.~\ref{fig:333_L_2}. DSU-Net demonstrates greater adaptability to these mutations than U-Net and STU-Net, which do not respond to mutations. Concretely, DSU-Net has better effect on FB when dealing with large vertical changes. At the same time, from the traces 125-145 in Fig.~\ref{fig:333_L_2}, it can be observed that DSU-Net is significantly better than other models in its ability to process fine, fragmented tubular features. 

The analysis of the visualized results above indicates that the benefits of DSU-Net are more noticeable in datasets with low SNRs. DSU-Net can effectively utilize the geometric characteristics of seismic shot gathers even in datasets with complex geological features. This demonstrates the adaptability of DSU-Net across various geological conditions, indicating that the model possesses superior generalization capabilities.
\begin{figure*}[!ht]
    \centering
    \subfloat[Direction (HR@1px)]{\includegraphics[width=0.29\textwidth]{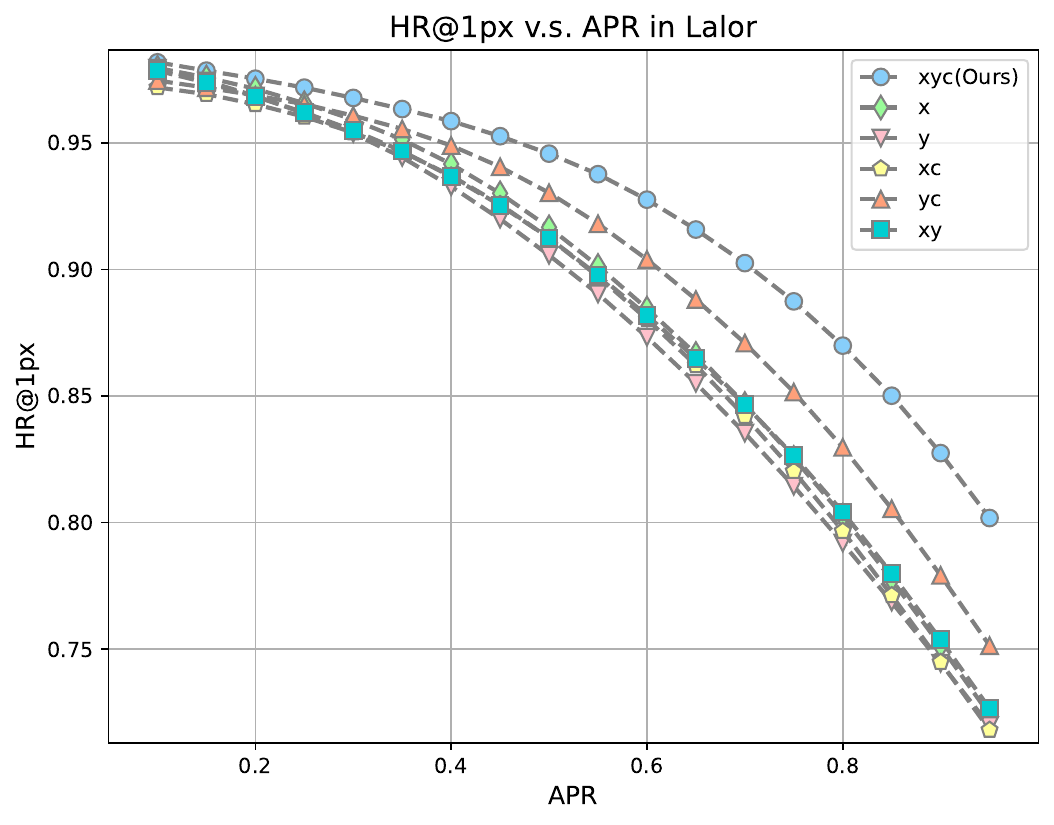}\label{fig:34_direction1}}
    \subfloat[Direction (HR@3px)]{\includegraphics[width=0.29\textwidth]{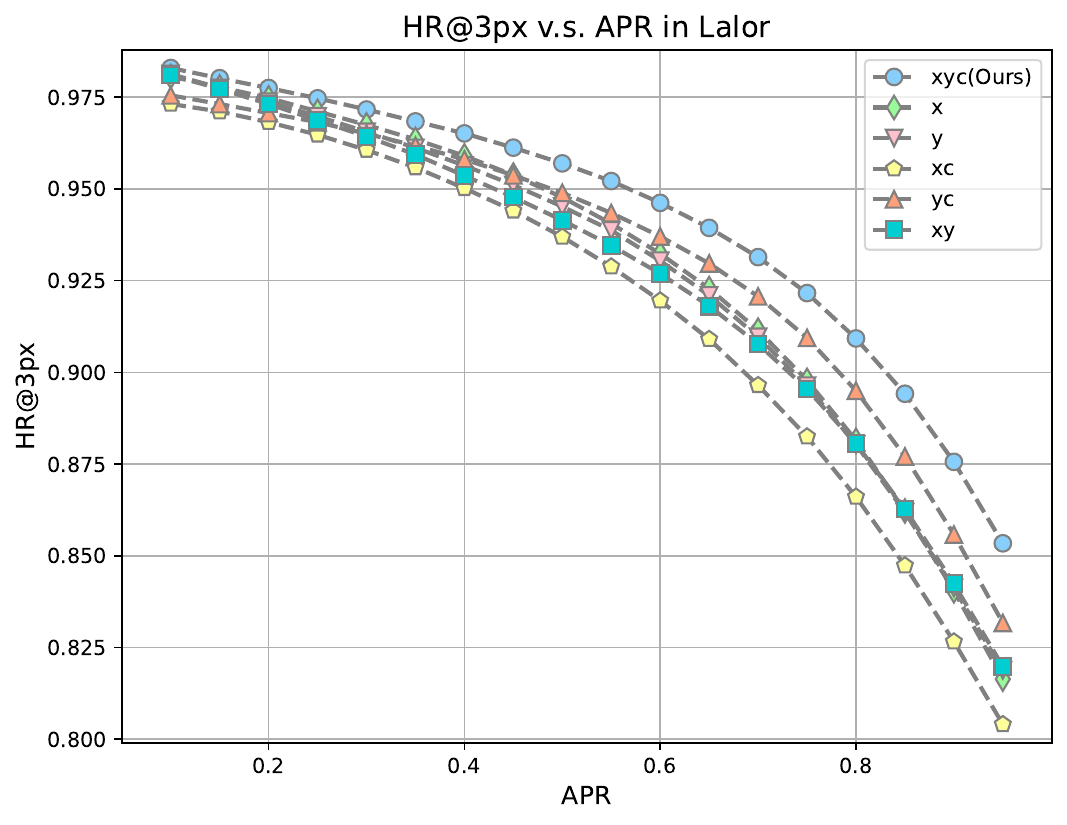}\label{fig:34_direction2}}
    \subfloat[Direction (MAE)]{\includegraphics[width=0.29\textwidth]{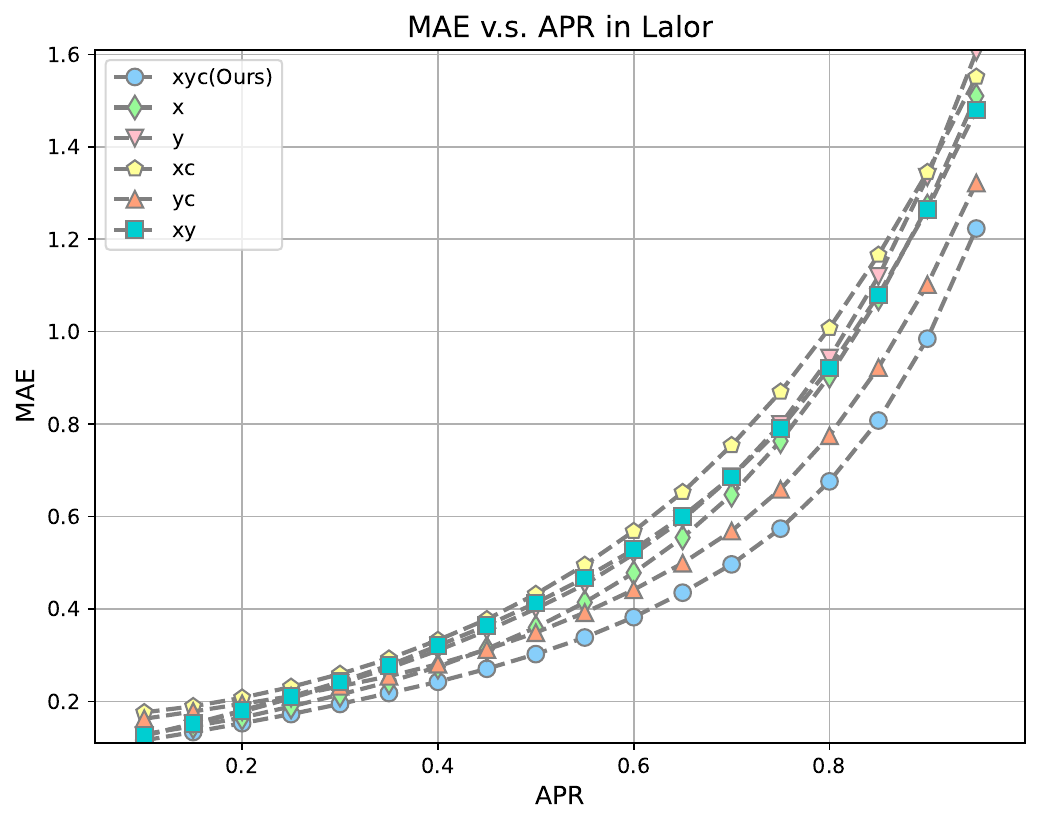}\label{fig:34_direction3}}\\
    \subfloat[Extension Scope (HR@1px)]{\includegraphics[width=0.29\textwidth]{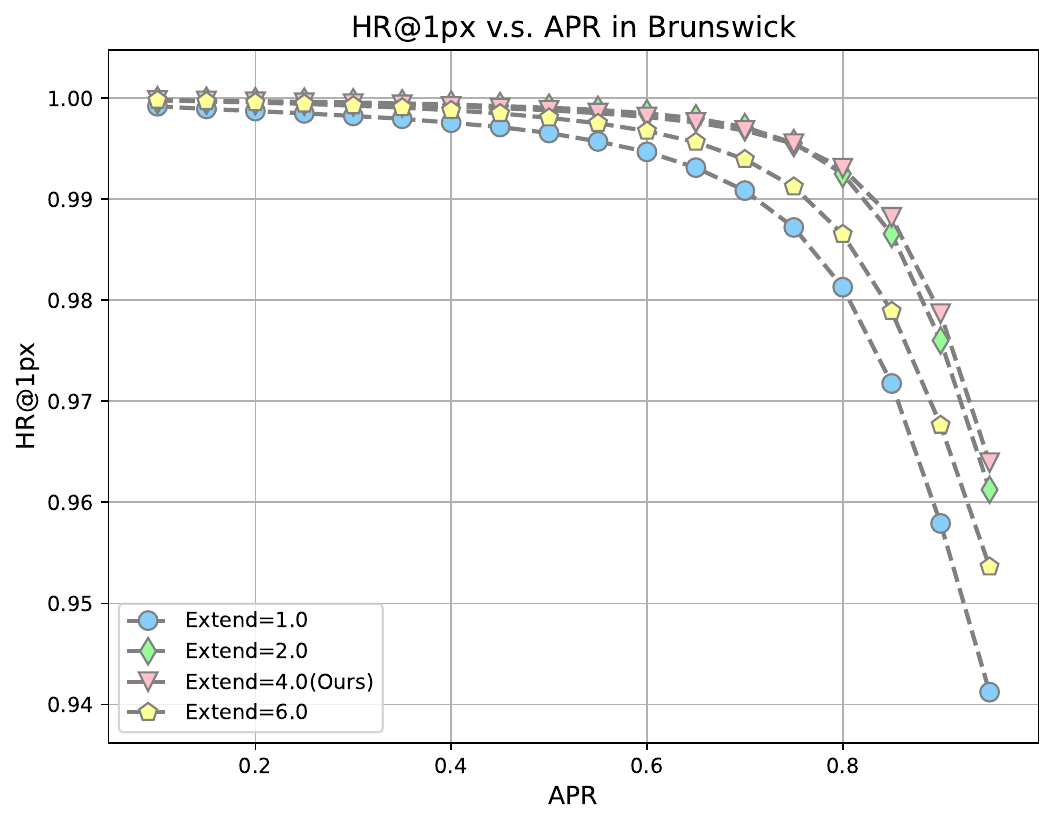}\label{fig:34_extention1}}
    \subfloat[Extension Scope (HR@3px)]{\includegraphics[width=0.29\textwidth]{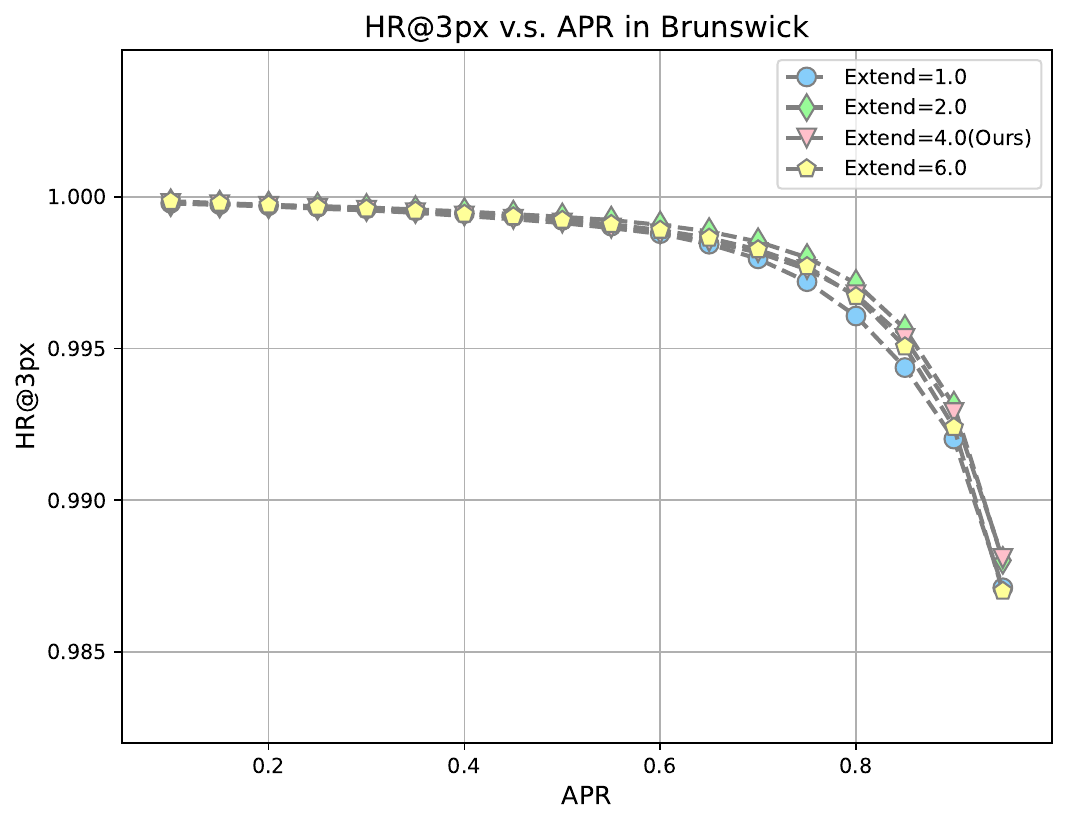}\label{fig:34_extention2}}
    \subfloat[Extension Scope (MAE)]{\includegraphics[width=0.29\textwidth]{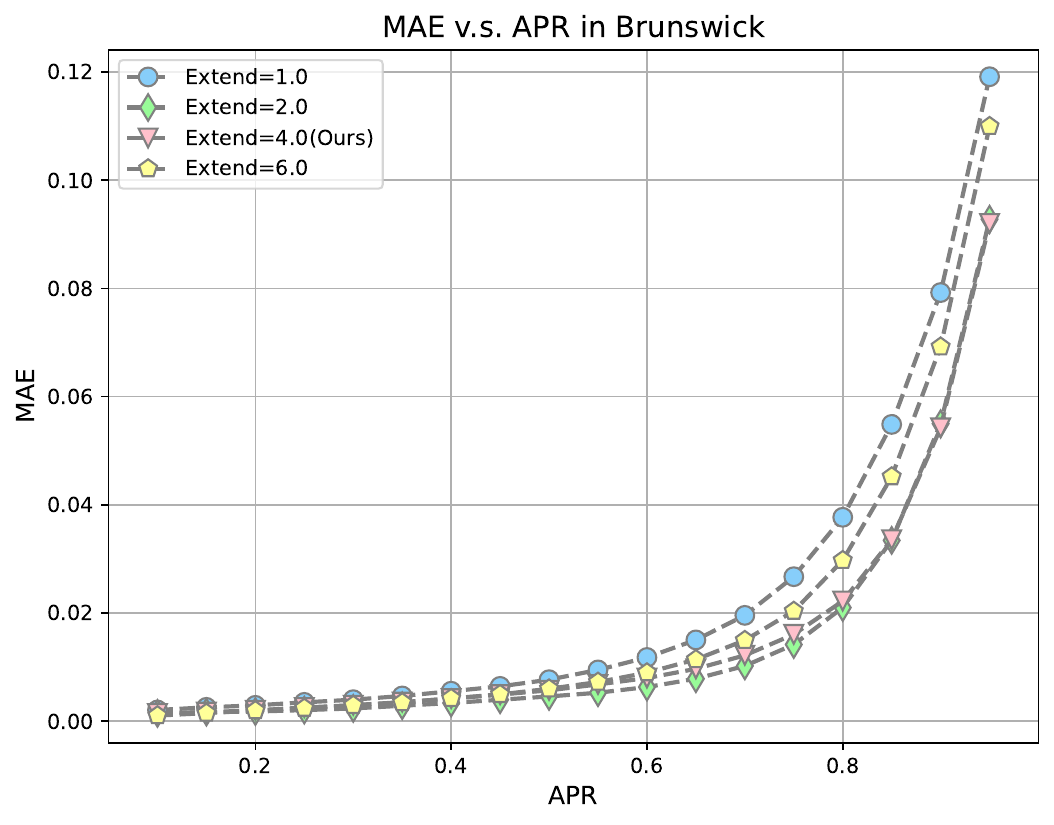}\label{fig:34_extention3}}\\
    \subfloat[Kernel Size (HR@1px)]{\includegraphics[width=0.29\textwidth]{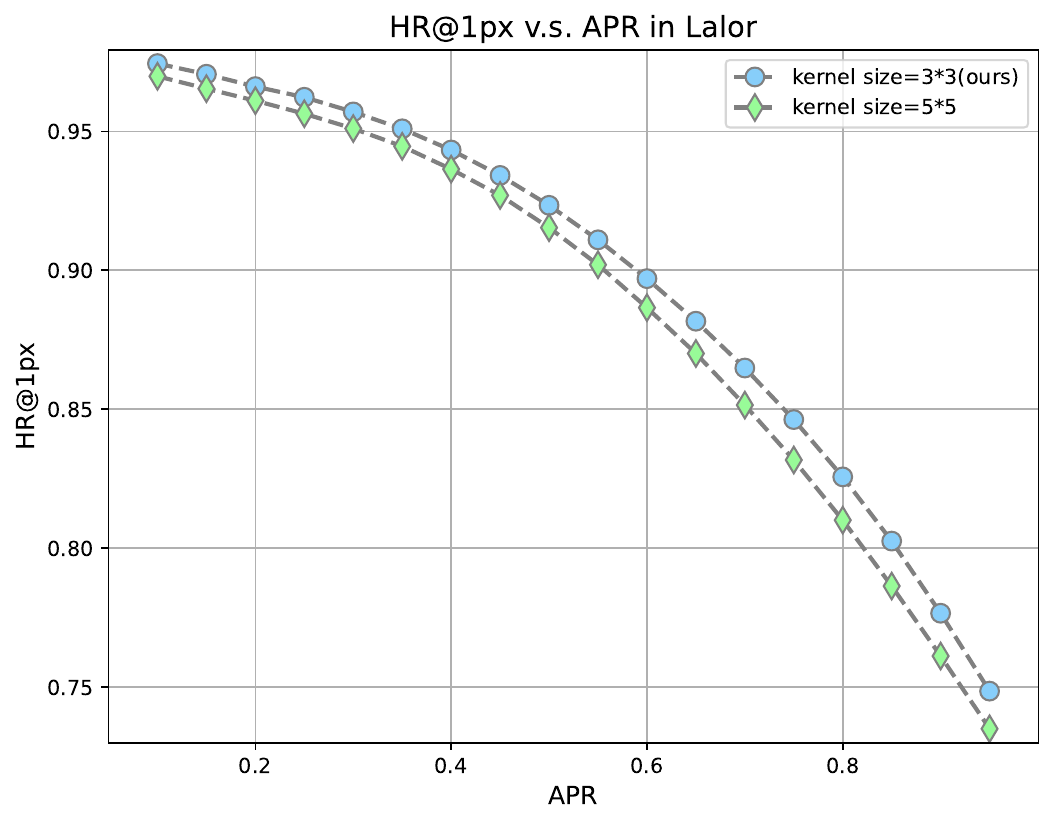}\label{fig:34_kernel1}}
    \subfloat[Kernel Size (HR@3px)]{\includegraphics[width=0.29\textwidth]{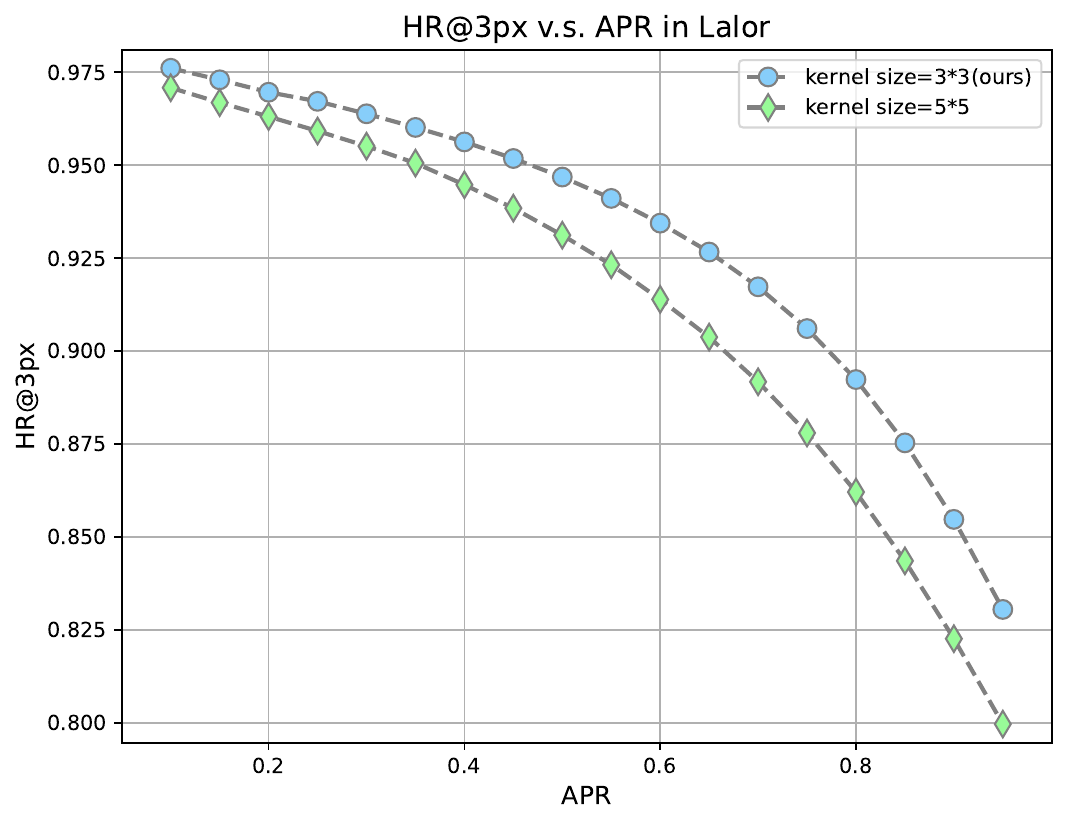}\label{fig:34_kernel2}}
    \subfloat[Kernel Size (MAE)]{\includegraphics[width=0.29\textwidth]{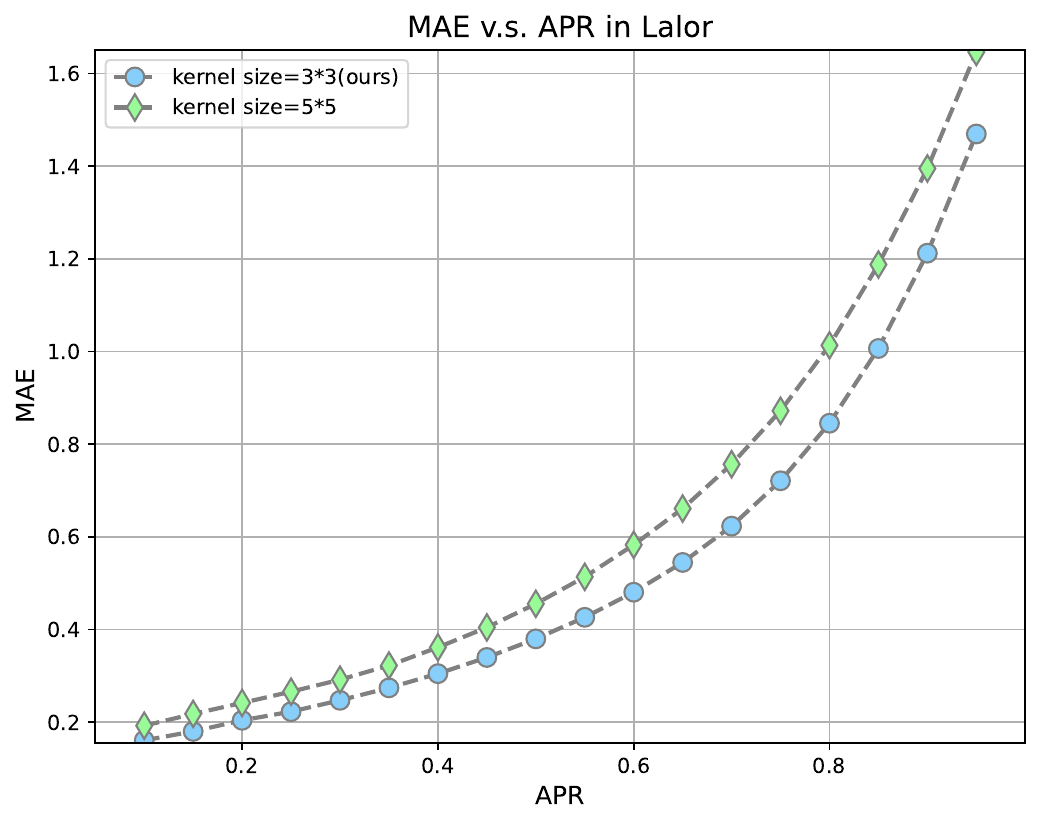}\label{fig:34_kernel3}}\\
    \subfloat[Data Augmentation (HR@1px)]{\includegraphics[width=0.29\textwidth]{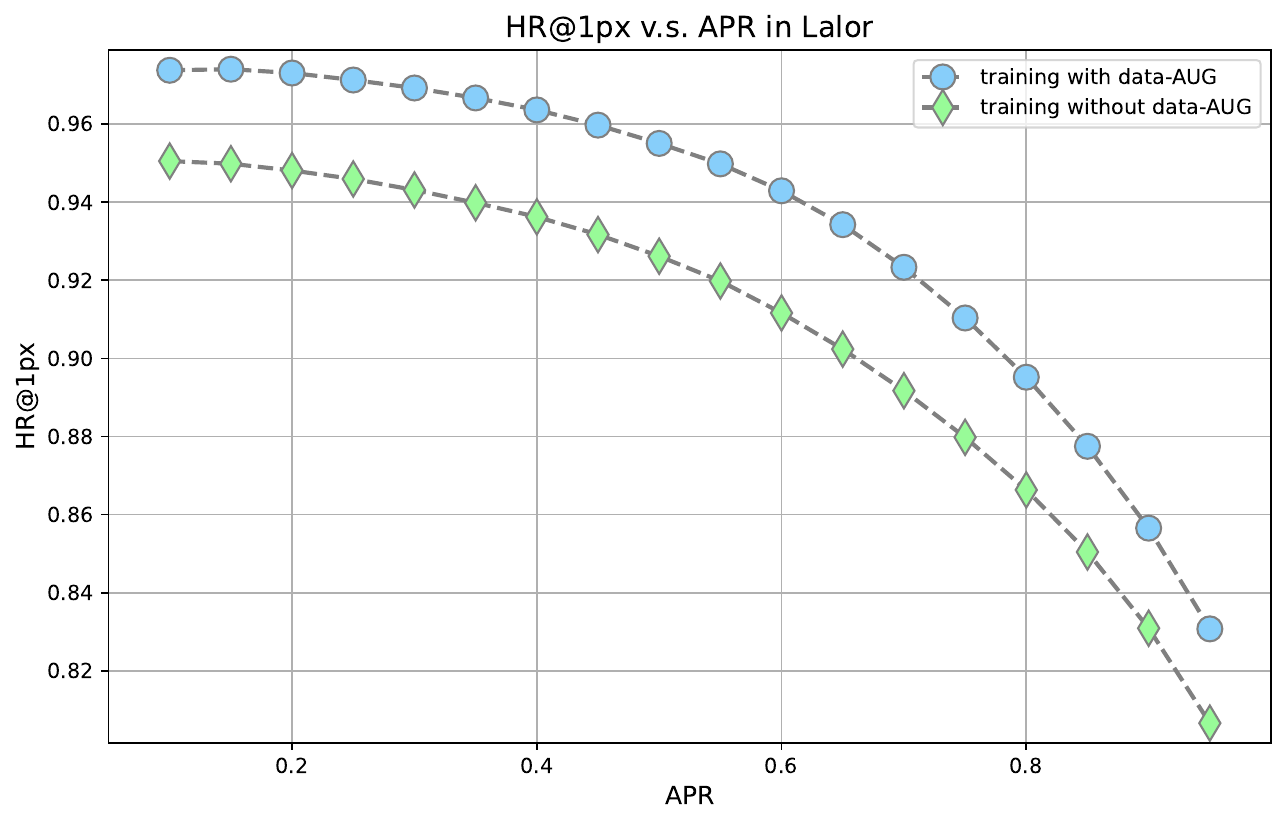}\label{fig:34_DA1}}
    \subfloat[Data Augmentation (HR@3px)]{\includegraphics[width=0.29\textwidth]{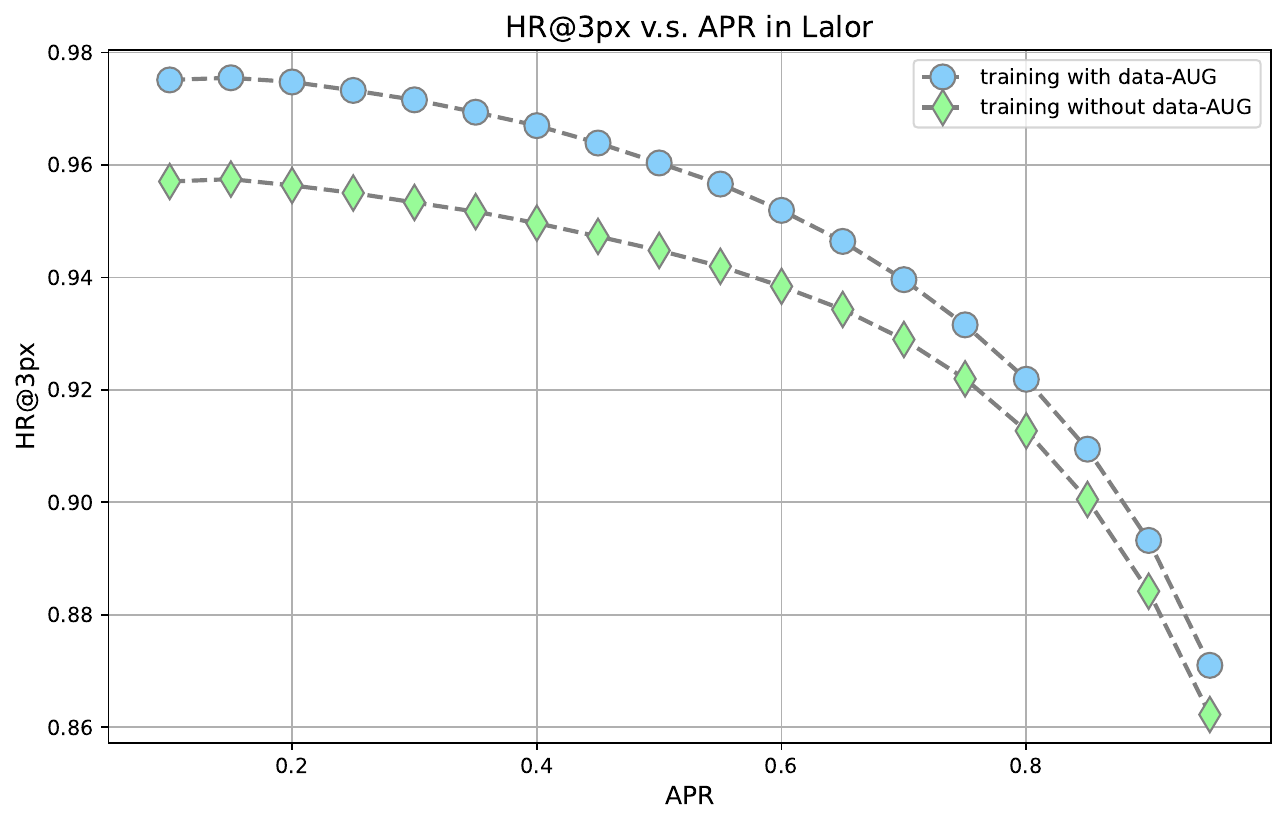}\label{fig:34_DA3}}
    \subfloat[Data Augmentation (MAE)]{\includegraphics[width=0.29\textwidth]{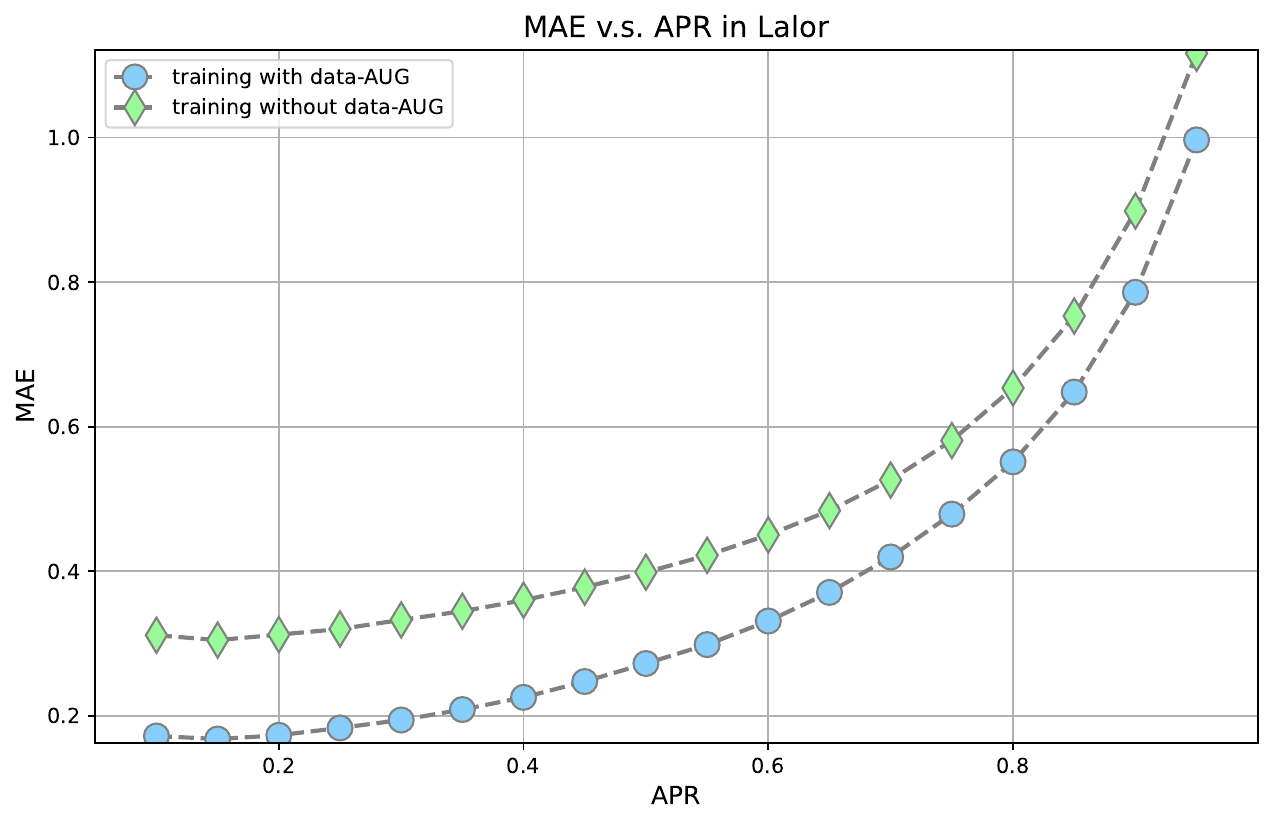}\label{fig:DA3}}
    \caption{Test results in the ablation studies.}
    \label{fig:ablation}
\end{figure*}

\subsection{Ablation Study}
\subsubsection{Model-Aspect Ablation}
To validate the rationality of the structure of our DSU-Net, we conduct ablation experiments on the multi-view feature maps of the DSConv module, as well as two hyper-parameters in the DSConv layer: the extension scope ($s_e$ in Eq.~\ref{eq:ds_conv_x}) and the kernel size.

On the one hand, we investigate the multi-view feature maps of our DSConv module. When using the DSConv to capture surface features, we implement a multi-view feature fusion strategy. In the first layer of the DSU-Net, during each feature extraction process, three convolution operations are executed on the input data: a traditional convolution operation, a DSConv operation along the horizontal direction (x-axis), and a DSConv operation along the vertical direction (y-axis). Subsequently, the generated three feature maps from these convolutions are combined as shown in Fig.~\ref{fig:DSUNet}b. 
Concretely, we design the other five methods to extract shallow features: (i) extracting feature maps from the x-axis direction only; (ii) extracting feature maps from the y-axis direction only; (iii) fusing the x-axis direction and ordinary convolutional feature maps; (iv) fusing the y-axis direction and ordinary convolutional feature maps; and (v) fusing the x-axis direction and fusing the y-axis direction feature map. 
We test these five models on Fold \#4 and compare the MAE, HR@1px, and HR@3px under various APRs. We plot a scatter chart with various APRs on the x-axis and MAE, HR@1px, and HR@3px on the y-axis as shown in Fig.~\ref{fig:34_direction1}-\ref{fig:34_direction3}. The experimental results shows that the performance of our current choice outperforms the other five models on the three metrics for every APR, verifying that the both features extracted by the DSConv modules along the x-axis and y-axis can significantly enhance the picking accuracy. 

On the other hand, we conduct a hyperparameter search for the extension scopes and the kernel sizes of the DSConv layer. The extension scope is used to control the expansion range of the coordinate map. To consider pixels further away when generating the coordinate map and thus create a more comprehensive map, the offset needs to be multiplied by the extension scope. If the expansion range is too small, it may limit the receptive field of the DSConv. Conversely, if the deformation range is too large, it may weaken the ability of the DSConv to extract local fine features, thereby affecting the overall performance of the DSU-Net. The kernel size represents the size of the convolution kernel. An excessively large convolution kernel leads to higher computational costs. It may reduce attention to certain local features, while an excessively small kernel may have a smaller receptive field, potentially resulting in incomplete feature extraction. In our experiment, the candidate values for the extension scope are 1.0, 2.0, 4.0, and 6.0.
Subsequently, the models with different extension scopes are randomly initialized on four datasets with five random seeds. Then, we plot the scatter chart with APR as the abscissa and MAE, HR@1px, and HR@3px as the ordinate. The representative results are shown in Fig.~\ref{fig:34_extention1}-\ref{fig:34_extention3}. It can be seen that when the extended scope is 4.0, the HR@1px and HR@3px of DSU-Net are the highest among all APRs. This suggests that when the extension scope is 4.0, the accuracy of the DSU-Net in picking results is the highest. In other words, the DSConv demonstrates the strongest ability to extract local small features when the extended scope is 4.0. In addition, the MAE performance is similar to that of the model with an extension scope of 2.0 on different APRs, and both are lower than the other two models. This suggests that when the extension scope of the DSConv module in the DSU-Net is 4.0 or 2.0, the receptive field becomes more concentrated on the features of interest after training.
Consequently, the relevant features are extracted more comprehensively, resulting in the smallest total error on the dataset. Therefore, an extension scope parameter of 4.0 for a serpentine convolution is optimal. 
We then adjust the kernel size to $3\times3$ and $5\times5$, respectively, and compare the impact of the DSConv module on the performance of the DSConv layer under different kernel sizes. MAE, HR@1px, and HR@3px are still selected as evaluation indicators. We also draw a scatter chart with different APRs as the abscissa and the above three indicators as the ordinate, and the representative results are shown in Fig.~\ref{fig:34_kernel1}-\ref{fig:34_kernel3}. The HR@1px and HR@3px of DSU-Net with a kernel size of $3\times3$ is significantly higher than that of DSU-Net with a kernel size of $5\times5$, and the MAE is considerably lower at all APRs. This demonstrates that when the kernel size of the DSConv layer is $3\times3$, it can accurately extract the initial features. However, with a kernel size of $5\times5$, the DSConv layer may overlook local  features due to the large receptive field. Consequently, it may fail to capture fine-grained local features, leading to a performance decline. Thus, in the DSU-Net, we use the DSConv layers with kernel size of $3\times3$ to enhance feature extraction capabilities.

\subsubsection{Optimization-Aspect Ablation}
To verify the impact of data augmentation on training results, we conduct an ablation experiment to determine whether data augmentation is performed during the training process. We perform two training processes: the training with data augmentation (training with data-AUG) and the training without data augmentation (training without data-AUG) on the Fold \#4, and compare MAE, HR@1px and HR@3px of two trained models respectively under different APRs. We use the testing results to draw a scatter chart for APRs and the above three metrics as shown in Fig.~\ref{fig:34_DA1}-\ref{fig:34_DA3}. The experimental results show that the training with data augmentation performs better than the training without data augmentation on the above three metrics in each APR, indicating that the proposed data augmentation method can effectively help the model recognize a wider data distribution. 

\subsection{Robustness Study}
\begin{figure}[!ht]
    \centering
    \subfloat[HR@1px]{\includegraphics[width=0.24\textwidth]{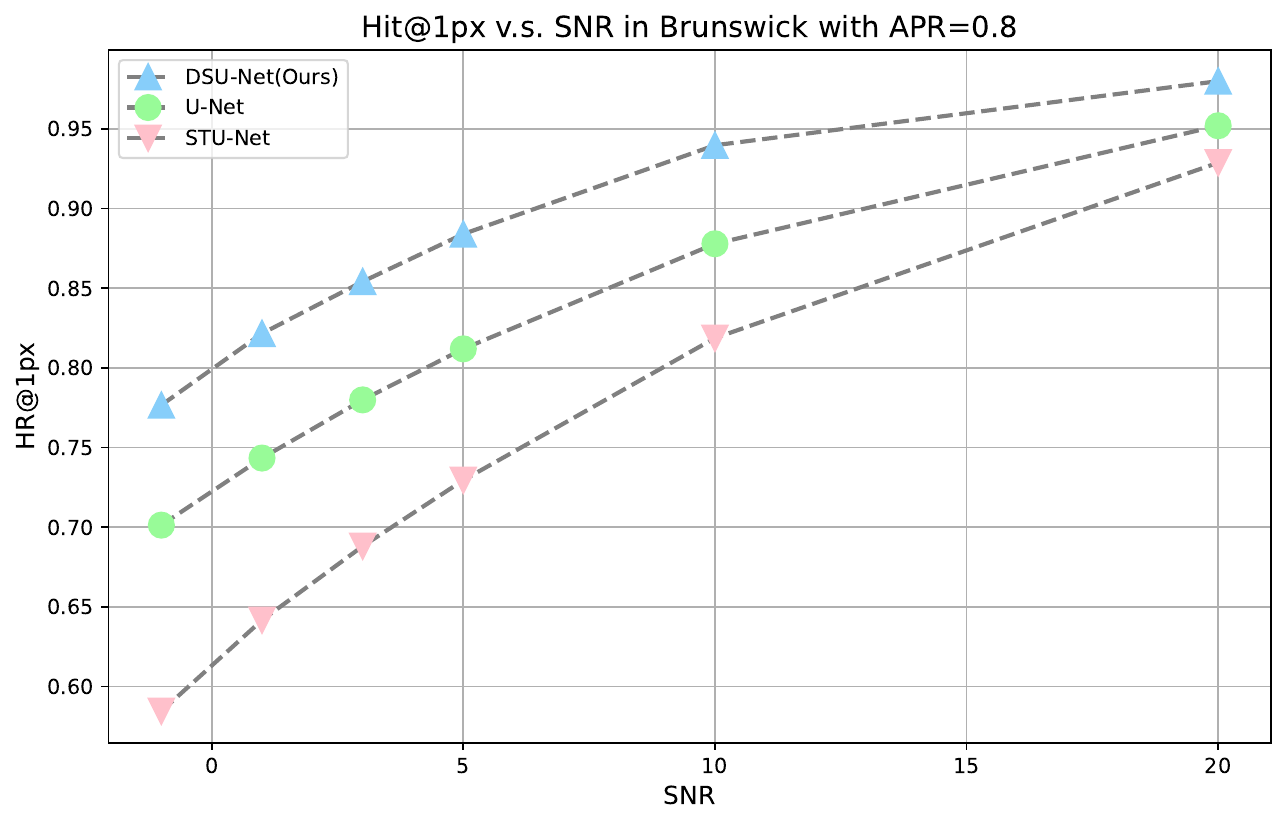}\label{fig:35_robust1}}
    \subfloat[HR@5px]{\includegraphics[width=0.24\textwidth]{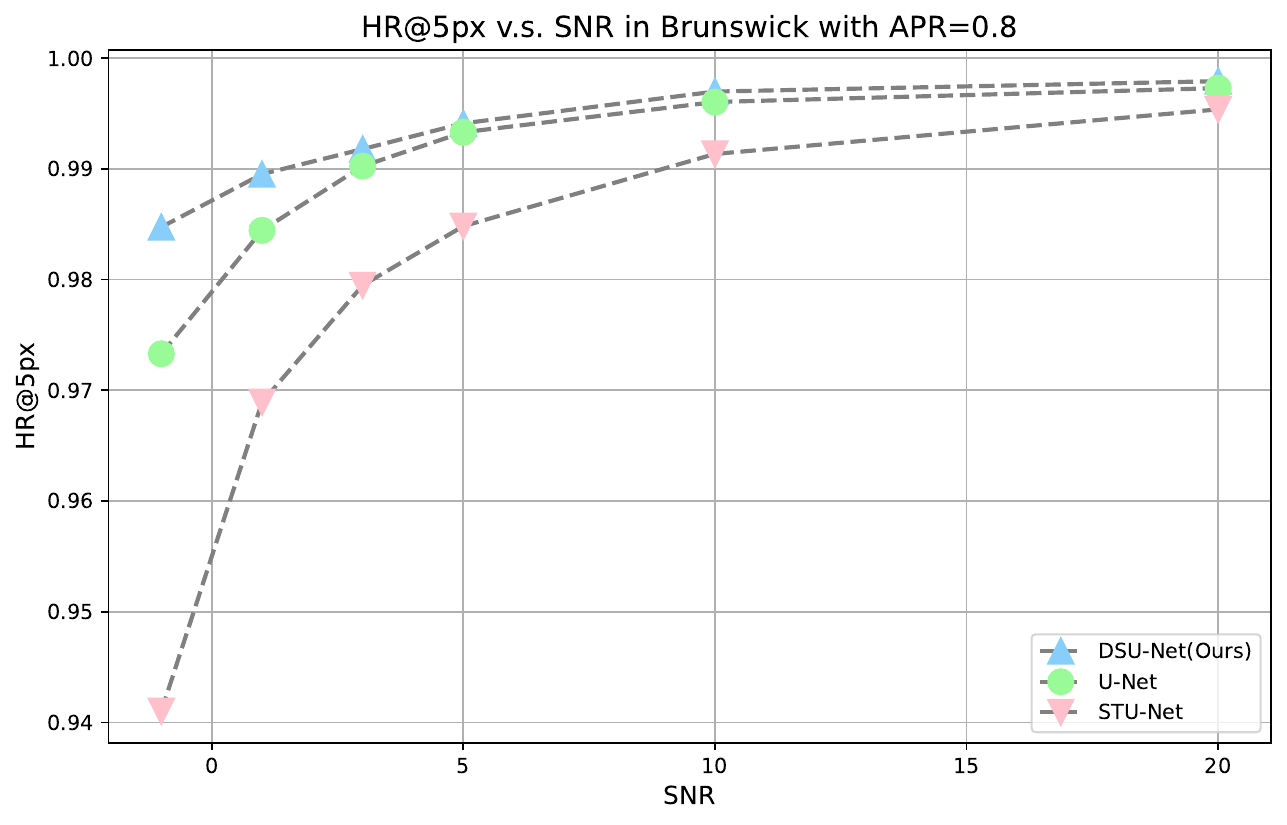}\label{fig:35_robust2}}\\ 
    \subfloat[MAE]{\includegraphics[width=0.24\textwidth]{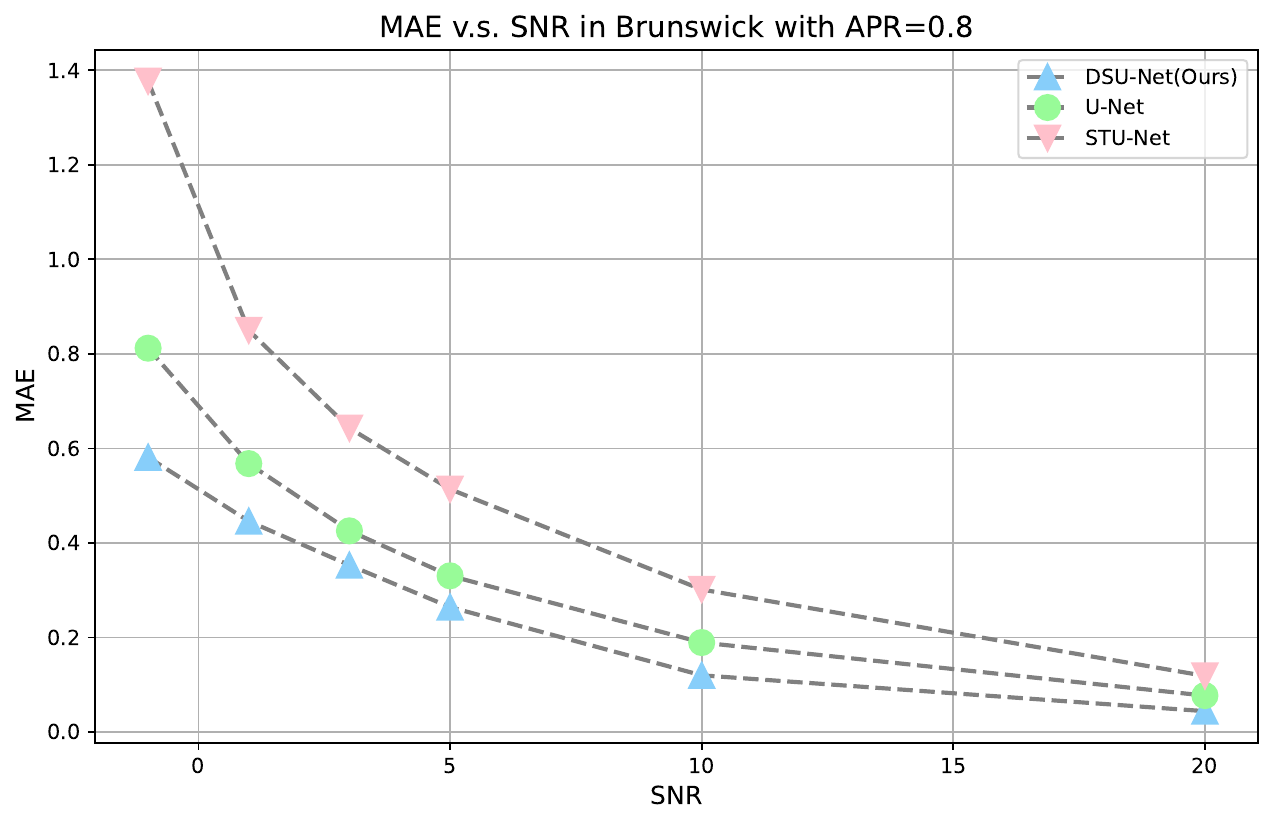}\label{fig:35_robust3}}
    \caption{Test results of the robustness test.}
    \label{fig:robust}
\end{figure}

To test the robustness of the DSU-Net, Gaussian noises with specific SNRs are added to the testing set, and the performances of the DSU-Net, the benchmark method (U-Net), and the STU-Net are compared on the testing set under each SNR. Since the Brunswick dataset has a high SNR, we assume that the Brunswick dataset is pure, meaning no noise is present. Thus, we can generate a polluted dataset with a desired constant SNR by injecting the Gaussian noises. The SNR of traces is given as follows:
\begin{equation}
    \text{SNR}=10\times \log_{10}\frac{\sigma^2_s}{\sigma^2_n},
 \label{eq:snr}
\end{equation}
where $\sigma^2_s$ and $\sigma^2_n$ represent the variances of the signal and the Gaussian noises, respectively. Therefore, we can adjust the Gaussian noise by changing the SNR by the following equation:
\begin{equation}
    \sigma^2_n = \sigma^2_s / 10^{\text{SNR}^*/10},
 \label{eq:snr2}
\end{equation}
where $\text{SNR}^*$ is a constant SNR and $\sigma^2_s$ can be calculated from the signal value on each gather. In our robustness experiment, we set the value of SNR* from small to large, i.e., -1,1,3,5,10,20 and generate six testing sets with the Gaussian noise of different SNRs. Then, we apply the trained DSU-Net model, the benchmark method, and the STU-Net on each polluted Brunswick dataset. We use HR@1px, HR@5px, and MAE as the evaluation metrics. With APR=0.8, the evaluation metrics of the DSU-Net, the benchmark method, and the STU-Net on datasets with different SNRs are indicated in a line chart, shown in Fig.~\ref{fig:robust}. HR@1px and HR@5px of the DSU-Net model are higher than the benchmark method and the STU-Net under different SNRs. Additionally, MAE is smaller than the benchmark method and the STU-Net under different SNR conditions.
Moreover, compared with the benchmark method and the STU-Net, when the SNR decreases, HR@1px and HR@5px of the DSU-Net decrease more slowly, and MAE rises more slowly. Therefore, we can conclude that the DSU-Net is more robust than the benchmark method and the STU-Net in the case of low SNRs. 

\section{Conclusion}
This paper proposes a novel segmentation network for FB picking called DSU-Net. After analyzing various experiments, three conclusions can be inferred. (1) Our proposed DSU-Net can pick FB accurately and stably and still has stable picking in the traces with continuous jumps of FBs.
(2) Compared with other 2-D FB picking models based on network improvement, DSU-Net achieves a SOTA level.
(3) Through detailed ablation experiments, we verify that the current design of the DSConv module is optimal for FB picking, providing a new backbone layer for other FB-picking networks.
In future work, we will optimize the picking framework to utilize the proposed segmentation network better.

\section*{Acknowledgment}
The author would like to thank Mr. Pierre-Luc St-Charles from Applied Machine Learning Research Team Mila, Québec AI Institute for providing the open datasets. 
\ifCLASSOPTIONcaptionsoff
  \newpage
\fi

\bibliographystyle{IEEEtran}
\bibliography{reference}

\end{document}